\documentclass[runningheads]{llncs}
\usepackage[a4paper,left=3cm,right=3cm]{geometry}
\usepackage[T1]{fontenc}
\usepackage{CharisSIL}
\usepackage{graphicx}
\usepackage{amsfonts}
\usepackage{amsmath}
\usepackage{algorithm}
\usepackage{algpseudocode}
\usepackage{censor,caption}
\usepackage{multirow}
\usepackage{graphicx,verbatim}
\usepackage{color}
\usepackage{booktabs}
\usepackage{array}
\usepackage{longtable}
\usepackage{tabularx}
\usepackage{threeparttable}
\usepackage{caption}
\usepackage{xurl}
\usepackage{amssymb}
\usepackage{xcolor}
\usepackage[colorlinks=true,linkcolor=blue,citecolor=blue,urlcolor=blue]{hyperref}
\makeatletter
\AtBeginDocument{\let\biasorigin@doi\doi\renewcommand{\doi}[1]{\in@{https://doi.org/}{#1}\ifin@\href{\detokenize{#1}}{\url{#1}}\else\biasorigin@doi{#1}\fi}}
\renewcommand\@seccntformat[1]{\csname the#1\endcsname.\quad}
\renewcommand\subsection{\@startsection{subsection}{2}{\z@}{-18\p@ \@plus -4\p@ \@minus -4\p@}{\baselineskip}{\normalfont\normalsize\itshape\rightskip=\z@ \@plus 8em\pretolerance=10000 }}
\renewcommand\subsubsection{\@startsection{subsubsection}{3}{\z@}{-12\p@ \@plus -4\p@ \@minus -4\p@}{.1\p@}{\normalfont\normalsize\itshape\rightskip=\z@ \@plus 8em\pretolerance=10000 }}
\makeatother

\begin{document}

\title{Race and gender differences in medical imaging models arise at the shared decision threshold}

\author{
Mahshad Lotfinia\inst{1}$^{\ast}$ \and
Daniel Truhn\inst{2,3} \and
Andreas Maier\inst{1} \and
Soroosh Tayebi Arasteh\inst{2,3}
}
\institute{
Pattern Recognition Lab, Friedrich-Alexander-Universit\"at Erlangen-N\"urnberg, Erlangen, Germany \and
Lab for AI in Medicine, RWTH Aachen University, Aachen, Germany \and
Department of Diagnostic and Interventional Radiology, University Hospital RWTH Aachen, Aachen, Germany
}

\maketitle
\pagestyle{plain}
\thispagestyle{plain}
{\footnotesize
\noindent$^{\ast}$Correspondence to:
Mahshad Lotfinia (\email{mahshad.lotfinia@fau.de})
}

\begin{abstract}
Fairness evaluations compare the performance of a medical imaging model between patient groups. A group difference in how well the model ranks images requires a better model, whereas a difference in error rates at the decision threshold shared by all groups can be addressed at the threshold. In studies of single chest radiograph models, race groups differed in error rates and hardly in ranking. However, to our knowledge, this has not been tested across image encoders, findings, and hospitals. We compared the area under the receiver operating characteristic curve (AUROC), sensitivity, and false positive rate (FPR) between patient groups for 133 classification models from 13 public image encoders. The data comprised more than 700{,}000 images from 10 different public datasets of chest radiography, dermatology, and funduscopy. The tests resampled patients. Within each of the two hospitals with recorded race, no race difference in AUROC was statistically significant in any of the 117 chest radiograph models. By contrast, the race difference in FPR was significant in 107 models in one hospital, at a median of 12 more false positive calls per 100 radiographs without the finding, and in 54 models in the other, at a median of 6. The same held for 299 models from 22 encoders pretrained in this study with three objectives or race-balanced data and from a randomly initialized encoder. So within a hospital, the race groups differed significantly at the shared decision threshold and not in the ranking, whatever the encoder. Developers and hospitals could examine the decision threshold for each patient group before retraining a model to remove a group difference.
\end{abstract}

\section{Introduction}

A medical imaging model that performs worse for one group of patients can delay the diagnosis of these patients or send them to unnecessary examinations. For this reason, fairness evaluations measure the performance of a model separately for patient groups, such as female and male patients, age bands, or race groups~\cite{Chen2023Algorithmic}. Such group differences have been reported for chest radiographs~\cite{SeyyedKalantari2021Underdiagnosis}, dermatology images~\cite{Daneshjou2022DDI}, retinal images~\cite{Burlina2021Retinal}, and pathology slides~\cite{Vaidya2024Pathology}. Reporting guidelines now ask for the performance of a model in patient subgroups~\cite{Collins2024TRIPODAI,Lekadir2025FUTUREAI}.

\begin{figure*}[!tbp]
\centering
\includegraphics[width=\textwidth]{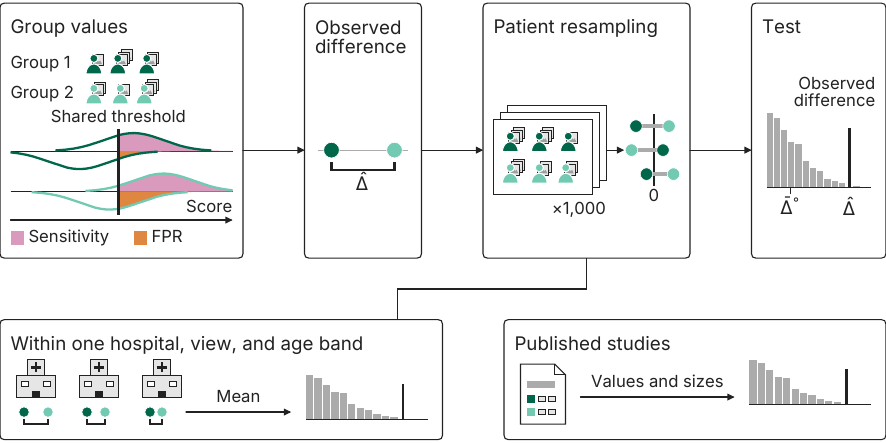}
\caption{Overview of the comparison of patient groups. Two groups whose scores differ by a shift have the same area under the receiver operating characteristic curve (AUROC) and differ in sensitivity and false positive rate (FPR) at a shared threshold. The observed difference between the groups, $\hat\Delta$, is compared with its distribution under equal group performance, which comes from 1{,}000 resamples of patients with all of their images. $\bar\Delta^{\circ}$ marks the difference expected by chance. The comparison is repeated within hospitals, radiograph views, and age bands. For published studies, it uses their reported group values and group sizes.}
\label{fig:overview}
\end{figure*}

A group difference can arise in two ways, and the two call for different responses. The first way is the ranking. The area under the receiver operating characteristic curve (AUROC) of a group measures how well the model distinguishes the images of this group with a finding from its images without the finding, at no particular threshold. A lower AUROC in one group means that the model itself distinguishes the images of this group less well, e.g., because the group is rare in the training data~\cite{Larrazabal2020Gender}. The second way is the decision threshold. A deployed model calls an image positive when its score is at least a threshold, which is the same for all groups. Sensitivity is the proportion of images with the finding that are called positive, and the false positive rate (FPR) is the proportion of images without the finding that are called positive. Two groups with the same AUROC differ in both rates when the scores of one group are higher overall (Fig.~\ref{fig:overview}). Such a difference can be changed at the threshold, without retraining the model~\cite{Hardt2016Equality}.

For chest radiographs, the second way has been observed in single models~\cite{Glocker2023Encoding,Sutariya2025Meval}. However, to our knowledge, it has not been tested whether this is the general case, i.e., whether it holds across image encoders, findings, and hospitals. An answer has to account for sampling variation and for the origin of the groups. The difference between the best and the worst group is above zero even when all groups perform equally, since each group value is estimated from a finite number of patients~\cite{Lum2022Debiasing}, and several images of one patient are not independent~\cite{Obuchowski1997Clustered}. Moreover, patient groups can differ in confounding factors~\cite{Bernhardt2022Potential,Mukherjee2022Confounding}. In a test set that combines hospitals, a difference between groups can be a difference between hospitals. Therefore, we address the following research questions:

\begin{description}
\item[RQ1:] Within one hospital, do medical imaging models differ between race and sex groups in ranking performance, or in sensitivity and FPR at the decision threshold shared by all groups?
\item[RQ2:] Do these group differences depend on the image encoder and on its pretraining objective and data?
\end{description}

Here, we compare the AUROC, the sensitivity, and the FPR between patient groups for 133 disease classification models. Each model is a logistic regression classifier on the frozen features of one of 13 public image encoders, with one decision threshold for all groups. The data comprise 702{,}206 images from 10 different public datasets of chest radiography, dermatology, and funduscopy. Every test resamples patients with all of their images, and a difference counts as statistically significant after correction for the number of tested models. For the chest radiograph models, we test the race and sex differences within each hospital, and within the radiograph views and age bands of a hospital. To answer RQ2, we add 299 models from 22 encoders pretrained in this study on chest radiographs, with three objectives, race-balanced data, and edited reports, and from a randomly initialized encoder. In supporting analyses, we apply the comparison to the group differences reported in nine published studies and measure the effect of counting images instead of patients. We show in which measures the groups differ, and we do not test a method that removes the differences.

Our main contributions are the following:

\begin{itemize}
\item We show that, within one hospital, the race groups of chest radiograph models differ in FPR at the shared decision threshold and not significantly in AUROC, and that the sex groups differ significantly in AUROC in a single dataset only. The group with the lowest FPR also tends to have the lowest sensitivity, so that the shared threshold is a different operating point in each group.
\item We show that this result holds for every tested encoder. It holds for medical and general encoders, for encoders pretrained in this study with three objectives, with race-balanced data, and with edited reports, and for a randomly initialized encoder.
\end{itemize}

\section{Related work}

\subsection{Ranking performance and error rates of patient groups}

Seyyed-Kalantari et al. reported that chest radiograph models called patients with a finding free of findings more often when the patients were female, younger, Black, Hispanic, or insured by Medicaid~\cite{SeyyedKalantari2021Underdiagnosis}. On MIMIC-CXR and CheXpert, Glocker et al. found similar receiver operating characteristic curves for the race groups of a DenseNet-121, with different sensitivity and FPR at the shared threshold~\cite{Glocker2023Encoding}. Sutariya and Petersen reported the same for one DenseNet-121 and the label No finding in MIMIC-CXR, with one radiograph per patient. The specificity differed significantly between the race groups, and the AUROC did not~\cite{Sutariya2025Meval}. Yang et al. evaluated subgroup performance in radiology, dermatology, and ophthalmology and found that models corrected to be fair within their original data distribution were not consistently fair on new test datasets~\cite{Yang2024Limits}. So the contrast between ranking performance and error rates has been observed in single convolutional models and for one or two labels. Whether it holds across current encoders, findings, and hospitals has not been tested, which is the first question of this study.

\subsection{Composition of the groups, acquisition, and mitigation}

A group difference depends on the patients and the images of each group. Bernhardt et al. argued that differences in disease prevalence and in label quality between groups complicate the study of underdiagnosis~\cite{Bernhardt2022Potential}. Mukherjee et al. argued that confounding factors must be accounted for before a difference is attributed to a group~\cite{Mukherjee2022Confounding}. Glocker et al. resampled test sets to account for race imbalance, age differences, and disease prevalence, which reduced some differences in the error rates~\cite{Glocker2023Encoding}. Lotter reduced an underdiagnosis difference between race groups with thresholds selected separately for the radiograph views~\cite{Lotter2024Acquisition}. Brown et al. developed shortcut testing, which uses multitask learning to assess whether shortcut learning contributes to subgroup differences~\cite{Brown2023ShorT}. FairREAD combines image features with demographic attributes and applies thresholds specific to each subgroup~\cite{Gao2026FairREAD}. These studies changed the test set, the model, or the threshold of one or a few models. None of them compared the groups within one hospital, radiograph view, and age band across encoders, and none varied the pretraining of the encoder, as this study does.

\subsection{Sampling variation and repeated images}

Sampling variation produces a positive difference between the best and the worst group even when all groups perform equally. Lum et al. studied this for measures of disparity, including the largest minus the smallest group value, and developed a corrected estimate of the variance between groups~\cite{Lum2022Debiasing}. The meval toolbox provides confidence intervals, studentized permutation tests, and a correction for multiple comparisons of subgroups~\cite{Sutariya2025Meval}. For several images per patient, Obuchowski extended nonparametric inference for the AUROC to clustered data and showed that tests reject too often when the correlation within patients is ignored~\cite{Obuchowski1997Clustered}. Genders et al. described methods for sensitivity and specificity that account for correlated observations within patients~\cite{Genders2012Clustered}. We resample patients with all of their images, so that every resample keeps this correlation. In an ablation, we measure how the test changes when images are counted instead.

\section{Materials and methods}

\subsection{Experimental design}

We compared the AUROC, the sensitivity, and the FPR of medical imaging models between patient groups. Each model is a logistic regression classifier fitted to the frozen features of one image encoder for one finding, with one decision threshold for all groups. The models detect 13 findings in chest radiographs, malignancy in dermatology images, and glaucoma in retinal images. The main panel comprises 117 chest radiograph models from 9 public encoders and 8 models from 8 public encoders in each of the other two modalities, i.e., 133 models from 13 different public encoders. A replication panel adds 299 chest radiograph models from 22 encoders pretrained in this study and from one randomly initialized encoder.

The study has five parts. (i) We tested the group differences over all test patients of each modality, resampling patients with all of their images. (ii) To answer RQ1, we tested the race and sex differences of the chest radiograph models in each hospital separately and within the radiograph views and age bands of the hospitals. We also compared the order of the groups by FPR with their order by sensitivity. (iii) To answer RQ2, we repeated these tests in the replication panel, whose encoders differ in pretraining objective and data. (iv) We applied a form of the test that needs only group values and group sizes to the group differences reported in published studies. (v) In an ablation, we replaced the resampling of patients by counts of images and measured how often each form of the test is significant for groups that differ only by chance.

\subsection{Datasets}

The study uses $n = 702{,}206$ images from 10 public datasets (Table~\ref{tab:data}). The chest radiographs ($n = 650{,}207$) come from MIMIC-CXR~\cite{Johnson2019MIMICCXR}, CheXpert~\cite{Irvin2019CheXpert}, NIH ChestX-ray14~\cite{Wang2017ChestXray14}, PadChest~\cite{Bustos2020PadChest}, VinDr-CXR~\cite{Nguyen2022VinDrCXR}, and VinDr-PCXR~\cite{Pham2022VinDrPCXR}. The dermatology images ($n = 41{,}999$) come from ISIC 2019~\cite{Tschandl2018HAM10000,Combalia2019BCN20000}, Fitzpatrick17k~\cite{Groh2021Fitzpatrick17k}, and the Diverse Dermatology Images (DDI) set~\cite{Daneshjou2022DDI}. The retinal images are scanning laser ophthalmoscopy images from Harvard-FairVision~\cite{Luo2024FairVision}, with one image per patient. The four chest radiograph datasets that release patient identifiers contain 219{,}827 patients. NIH ChestX-ray14, VinDr-CXR, VinDr-PCXR, and Harvard-FairVision come with released test sets. We split the other datasets into training, validation, and test sets by patient, by lesion in ISIC 2019, or by image where no patient identifier exists. Thus, images from one identified patient or lesion remain in the same split. The dataset-specific curation, released splits, and labels are described in Supplementary Note~\ref{snote:datasets}.

\begin{table}[!tbp]
\centering
\caption{Datasets and image encoders. VinDr-CXR, VinDr-PCXR, Fitzpatrick17k, and DDI release no patient identifier. Width is the dimension of the feature vector of an encoder. DDI, Diverse Dermatology Images; N/A, not applicable; ViT, vision transformer.}
\label{tab:data}
\setlength{\tabcolsep}{4pt}
\renewcommand{\arraystretch}{1.08}
{\footnotesize\hyphenpenalty=10000\exhyphenpenalty=10000
\begin{tabular*}{\textwidth}{@{\extracolsep{\fill}}>{\raggedright\arraybackslash}p{0.27\textwidth}rrr>{\raggedright\arraybackslash}p{0.19\textwidth}@{}}
\toprule
Dataset & Images & Test images & Test patients & Groups compared \\
\midrule
\multicolumn{5}{@{}l}{\textbf{Chest radiography}} \\
\midrule
MIMIC-CXR~\cite{Johnson2019MIMICCXR} & 243{,}345 & 44{,}636 & 11{,}179 & Race, sex, age, insurance \\
CheXpert~\cite{Irvin2019CheXpert} & 157{,}865 & 29{,}318 & 9{,}810 & Race, sex, age, insurance \\
NIH ChestX-ray14~\cite{Wang2017ChestXray14} & 112{,}120 & 25{,}596 & 2{,}797 & Sex, age \\
PadChest~\cite{Bustos2020PadChest} & 110{,}525 & 22{,}045 & 13{,}438 & Sex, age \\
VinDr-CXR~\cite{Nguyen2022VinDrCXR} & 18{,}000 & 3{,}000 & N/A & Sex, age \\
VinDr-PCXR~\cite{Pham2022VinDrPCXR} & 8{,}352 & 1{,}397 & N/A & None \\
\midrule
\multicolumn{5}{@{}l}{\textbf{Dermatology}} \\
\midrule
ISIC 2019~\cite{Tschandl2018HAM10000,Combalia2019BCN20000} & 25{,}331 & 7{,}636 & 4{,}179 lesions & Sex, age \\
Fitzpatrick17k~\cite{Groh2021Fitzpatrick17k} & 16{,}012 & 4{,}804 & N/A & Skin type \\
DDI~\cite{Daneshjou2022DDI} & 656 & 196 & N/A & Skin type \\
\midrule
\multicolumn{5}{@{}l}{\textbf{Funduscopy}} \\
\midrule
Harvard-FairVision~\cite{Luo2024FairVision} & 10{,}000 & 3{,}000 & 3{,}000 & Race, ethnicity, sex, age \\
\midrule
\multicolumn{5}{@{}l}{\textbf{Image encoders}} \\
\midrule
Encoder & \multicolumn{1}{l}{Objective} & \multicolumn{1}{l}{Domain} & Width & Modalities \\
\cmidrule(lr){1-5}
RAD-DINO~\cite{PerezGarcia2025RadDino} & \multicolumn{1}{l}{Self-supervised} & \multicolumn{1}{l}{Chest radiographs} & 768 & Chest radiography \\
BiomedCLIP~\cite{Zhang2023BiomedCLIP} & \multicolumn{1}{l}{Image-text} & \multicolumn{1}{l}{Biomedical} & 512 & Chest radiography \\
TorchXRayVision DenseNet-121~\cite{Cohen2022TorchXRayVision} & \multicolumn{1}{l}{Supervised} & \multicolumn{1}{l}{Chest radiographs} & 1{,}024 & Chest radiography \\
DINOv2 ViT-L~\cite{Oquab2024DINOv2} & \multicolumn{1}{l}{Self-supervised} & \multicolumn{1}{l}{General} & 1{,}024 & All 3 \\
DINOv3 ViT-S, ViT-B, ViT-L~\cite{Simeoni2025DINOv3} & \multicolumn{1}{l}{Self-supervised} & \multicolumn{1}{l}{General} & 384, 768, 1{,}024 & All 3 \\
CLIP ViT-L/14~\cite{Radford2021CLIP} & \multicolumn{1}{l}{Image-text} & \multicolumn{1}{l}{General} & 768 & All 3 \\
SigLIP 2 ViT-L~\cite{Tschannen2025SigLIP2} & \multicolumn{1}{l}{Image-text} & \multicolumn{1}{l}{General} & 1{,}024 & All 3 \\
MONET~\cite{Kim2024MONET} & \multicolumn{1}{l}{Image-text} & \multicolumn{1}{l}{Dermatology} & 768 & Dermatology \\
DermLIP ViT-B/16~\cite{Yan2025DermLIP} & \multicolumn{1}{l}{Image-text} & \multicolumn{1}{l}{Dermatology} & 512 & Dermatology \\
RETFound~\cite{Zhou2023RETFound} & \multicolumn{1}{l}{Self-supervised} & \multicolumn{1}{l}{Fundus} & 1{,}024 & Funduscopy \\
FLAIR~\cite{Silva2025FLAIR} & \multicolumn{1}{l}{Image-text} & \multicolumn{1}{l}{Fundus} & 512 & Funduscopy \\
\bottomrule
\end{tabular*}}
\end{table}

The 13 chest radiograph findings are atelectasis, cardiomegaly, consolidation, edema, enlarged cardiomediastinum, fracture, lung lesion, lung opacity, pleural effusion, pleural other, pneumonia, pneumothorax, and support devices. A finding labeled present counts as positive, and a finding labeled absent or not mentioned counts as negative. An image with an uncertain label for a finding is left out for that finding. MIMIC-CXR and CheXpert provide test labels for all 13 findings, NIH ChestX-ray14 and PadChest for 6 findings each, and VinDr-CXR for 10. So 117 models are tested in MIMIC-CXR and in CheXpert, 54 in NIH ChestX-ray14 and in PadChest, and 90 in VinDr-CXR.

The dermatology models classify a lesion as malignant or benign. The dermatology split uses a fraction of 0.6 for training, 0.1 for validation, and 0.3 for testing, at seed 42. The funduscopy models detect glaucoma.

On chest radiographs, we compared patients by recorded race, sex, age, and insurance. Race and insurance are available only for MIMIC-CXR and CheXpert. For MIMIC-CXR, these attributes come from MIMIC-IV version 3.1~\cite{Johnson2023MIMICIV}. A patient's race and insurance are the most frequent recorded values over their admissions. For CheXpert, race and insurance come from the race and insurance fields of CheXpert Plus~\cite{Chambon2024ChexpertPlus}. Race comparisons use White, Black, Asian, and Hispanic or Latino, and insurance comparisons use Medicare, Medicaid, and private insurance. Patients with other or missing values are excluded only from the corresponding attribute comparison. Sex is recorded as male or female. The age bands are younger than 40, 40 to 59, 60 to 79, and 80 years or older. In dermatology, we compared sex and age in ISIC 2019 and Fitzpatrick skin types I and II, III and IV, and V and VI in Fitzpatrick17k and DDI. In funduscopy, we compared White, Black, and Asian race groups, Hispanic and non-Hispanic ethnicity, sex, and age. Table~\ref{tab:groups} lists the numbers of test patients and images in each group. Each chest radiograph dataset comes from one hospital, except VinDr-CXR, which comes from two hospitals, and we treat each dataset as one hospital. Radiograph view is anteroposterior or posteroanterior, and VinDr-CXR and VinDr-PCXR record no view.

The race groups come from the two hospitals in different proportions. All 552 Hispanic or Latino test patients come from MIMIC-CXR, as do 1{,}567 of the 2{,}082 Black test patients. Of the 1{,}407 Asian test patients, 1{,}082 are from CheXpert. Therefore, we test the race differences in each hospital separately.

\begin{table}[!tbp]
\centering
\caption{Test patients and test images of the compared groups. The counts include test images that have no label for a finding. A lesion counts as a patient in ISIC 2019. An image counts as a patient in DDI, Fitzpatrick17k, and VinDr-CXR, which release no patient identifier. Harvard-FairVision has one image per patient. Age is given in years. A patient whose images fall in more than one age band is counted in each.}
\label{tab:groups}
\setlength{\tabcolsep}{4pt}
\renewcommand{\arraystretch}{1.08}
{\footnotesize
\begin{tabular*}{\textwidth}{@{\extracolsep{\fill}}llrr@{}}
\toprule
Attribute & Group & Test patients & Test images \\
\midrule
\multicolumn{4}{@{}l}{\textbf{Chest radiography}} \\
Race & White & 10{,}972 & 42{,}477 \\
 & Black & 2{,}082 & 8{,}446 \\
 & Asian & 1{,}407 & 4{,}368 \\
 & Hispanic or Latino & 552 & 2{,}180 \\
Sex & Female & 18{,}289 & 52{,}703 \\
 & Male & 19{,}552 & 67{,}350 \\
Age & Younger than 40 & 10{,}608 & 24{,}015 \\
 & 40 to 59 & 12{,}080 & 39{,}390 \\
 & 60 to 79 & 13{,}082 & 43{,}829 \\
 & 80 or older & 4{,}997 & 14{,}566 \\
Insurance & Medicare & 8{,}408 & 35{,}014 \\
 & Medicaid & 2{,}586 & 9{,}706 \\
 & Private insurance & 5{,}381 & 16{,}523 \\
\midrule
\multicolumn{4}{@{}l}{\textbf{Dermatology}} \\
Skin type & I and II & 2{,}394 & 2{,}394 \\
 & III and IV & 1{,}917 & 1{,}917 \\
 & V and VI & 689 & 689 \\
Sex & Female & 1{,}920 & 3{,}539 \\
 & Male & 2{,}159 & 3{,}994 \\
Age & Younger than 40 & 805 & 1{,}422 \\
 & 40 to 59 & 1{,}678 & 2{,}817 \\
 & 60 to 79 & 1{,}228 & 2{,}414 \\
 & 80 or older & 351 & 860 \\
\midrule
\multicolumn{4}{@{}l}{\textbf{Funduscopy}} \\
Race & White & 2{,}334 & 2{,}334 \\
 & Black & 422 & 422 \\
 & Asian & 244 & 244 \\
Ethnicity & Non-Hispanic & 2{,}896 & 2{,}896 \\
 & Hispanic & 104 & 104 \\
Sex & Female & 1{,}746 & 1{,}746 \\
 & Male & 1{,}254 & 1{,}254 \\
Age & Younger than 40 & 331 & 331 \\
 & 40 to 59 & 847 & 847 \\
 & 60 to 79 & 1{,}554 & 1{,}554 \\
 & 80 or older & 268 & 268 \\
\bottomrule
\end{tabular*}}
\end{table}

\subsection{Encoders and classification models}

\subsubsection{Public encoders and feature extraction}

Table~\ref{tab:data} identifies each encoder, its pretraining objective and domain, and its feature dimension. The chest radiograph models use the three medical encoders and the six general encoders. The dermatology and funduscopy models use the general encoders and two encoders of the respective modality. DINOv2 and DINOv3 have previously been evaluated for chest radiograph classification~\cite{TayebiArasteh2024Enhancing,TayebiArasteh2026Resolution}. RAD-DINO and the TorchXRayVision DenseNet-121 were pretrained on data that include MIMIC-CXR, CheXpert, NIH ChestX-ray14, and PadChest, RAD-DINO without labels and the DenseNet-121 with labels. The released list of the pretraining images of RAD-DINO contains 114{,}625 of our 125{,}992 chest radiograph test images, i.e., 40{,}828 from MIMIC-CXR, all 29{,}318 from CheXpert, all 25{,}596 from NIH ChestX-ray14, and 18{,}883 from PadChest. We matched the images by their file name and, for CheXpert, by patient, study, and view. Therefore, we repeated the FDR correction of the tests within strata without these two encoders.

Each encoder maps an image to one feature vector. We use the last-layer class token for the self-supervised encoders, the pooled projection for the language-supervised encoders, and the pooled feature before the classification layer for DenseNet-121. Every image undergoes a Lanczos resize to $224 \times 224$ pixels before the encoder's released preprocessing. RAD-DINO then resizes the image to $518 \times 518$ pixels, and SigLIP 2 to $512 \times 512$ pixels. FLAIR receives the stored image with intensities scaled to $[0, 1]$. Feature extraction uses batches of 64 images. DenseNet-121 and FLAIR use single precision. The other encoders use 16-bit precision, with bfloat16 or single precision when needed to obtain finite features. All extracted features are stored in single precision and scaled to unit $L_2$ norm.

\subsubsection{Replication encoders}

The replication panel includes a ViT-S/16~\cite{Dosovitskiy2021ViT} with the default random initialization of timm, drawn without a fixed seed. Its 22 pretrained encoders start from DINOv3 ViT-S/16 or ViT-B/16 and use the $n = 180{,}300$ radiographs of the MIMIC-CXR training set or a race-balanced subset. Training lasts 50 epochs on $224 \times 224$ pixel inputs in batches of 256 images. It uses AdamW~\cite{Loshchilov2019AdamW} with a peak learning rate of $1 \times 10^{-4}$, weight decay of 0.05, linear warmup followed by cosine decay, and bfloat16 autocasting. The self-supervised objective uses a contrastive loss between two augmented views of each radiograph at temperature 0.1~\cite{Chen2020SimCLR}. The label-supervised objective uses binary cross-entropy over the 13 findings. The image-report objective uses a contrastive loss at temperature 0.07~\cite{Radford2021CLIP}. Its report encoder is Bio\_ClinicalBERT~\cite{Alsentzer2019ClinicalBERT}. At seed 0, 12 runs cross the 3 objectives, 2 backbones, and 2 data compositions. Another 4 image-report runs at seed 0 use reports with demographic terms removed or with a sentence naming race, sex, and age band added. The remaining 6 runs repeat the 3 objectives with the ViT-S backbone on the natural data at seeds 1 and 2. The paired training comparisons are reported in Supplementary Note~\ref{snote:replication}.

The image-report runs update weights after every batch, warm up for the first 10\% of steps, and reduce the learning rate to zero at the last step. The self-supervised and label-supervised runs use 512 images per optimizer step by accumulating gradients over two batches. The self-supervised negatives come from one batch of 256 images. For these two objectives, the schedule uses the number of batches, which is twice the number of optimizer steps. Therefore, warmup lasts for the first 20\% of optimizer steps, and the final learning rate is 59\% of its peak.

The augmentation starts with a random resized crop covering 20\% to 100\% of the image for the self-supervised objective and 50\% to 100\% for the other two objectives. A random horizontal flip follows with probability 0.5. The self-supervised objective also applies color jitter with brightness and contrast 0.4 at probability 0.5. Its loss is symmetric, and the negatives of each image come from the other view only. The label-supervised loss is a masked mean binary cross-entropy over the 13 findings. The image-report runs pair each radiograph with the full text of its MIMIC-CXR report, truncated at 256 tokens, and train the text encoder at the same learning rate. The text representation is its last-layer class token. For the self-supervised objective, a 2-layer projection head of width 2{,}048 maps the backbone output to 256 dimensions. For the image-report objective, a linear projection maps each encoder output to 256 dimensions. The projection heads are discarded at feature extraction.

For the race-balanced subset, we sampled 5{,}576 radiographs from each category at seed 42 to match the smallest category's size. The five categories are Asian, Black, Hispanic or Latino, White, and Other. The Other category is included in pretraining but excluded from the race comparisons. One report condition uses case-insensitive whole-word matching to remove 14 sex and race terms and every age written in digits. The other report condition adds one sentence that names the race, the sex, and the age band of the patient at the start of each report. For feature extraction, each stored image is resized to $256 \times 256$ pixels, center-cropped to $224 \times 224$ pixels, and normalized with the ImageNet channel statistics. The feature vector is the class token of the last layer.

\subsubsection{Classification and threshold selection}

For each encoder and finding, we fitted one logistic regression classifier in scikit-learn~\cite{Pedregosa2011Sklearn} using the labeled training images pooled within the modality. When more than 100{,}000 training images were available, we sampled 100{,}000 without replacement at seed 0. Feature standardization was fitted on those training images and applied unchanged to validation and test images. The classifier uses $L_2$ regularization with $C = 1$, the L-BFGS solver, at most 1{,}000 iterations, and class weights inverse to the class frequencies. Every attribute comparison uses the same fitted classifier.

For each classifier, the decision threshold is the 20th percentile of the scores of positive validation images, calculated by linear quantile interpolation. This corresponds to a validation sensitivity of 0.80. The threshold is fixed before test evaluation and shared by all groups, hospitals, and strata of that model. Neither the classifier nor its threshold is refitted when the test patients are resampled.

\subsection{Group differences and the difference expected by chance}

For a model and an attribute, let $\hat m_j$ be the AUROC, the sensitivity, or the FPR of group $j$ on the test set. All three measures count images. The AUROC compares the scores of the images with and without the finding, and a pair of tied scores counts 0.5. Sensitivity is the fraction of the images with the finding that are called positive, and FPR is the fraction of the images without the finding that are called positive. An image is called positive when its score is at least the threshold of the model. A group is eligible when it has at least 20 labeled test images, with at least one image with and one without the finding. With $k \ge 2$ eligible groups, the group difference is
\begin{equation}\label{eq:obs}
\hat\Delta = \max_{j} \hat m_j - \min_{j} \hat m_j .
\end{equation}
The difference is 0 only when all groups have the same value, and it cannot be negative. It does not show which group has the higher value, so we report the groups with the highest and the lowest value separately. A higher AUROC, a higher sensitivity, and a lower FPR are better. For example, an FPR difference of 0.10 means 10 more false positive calls per 100 images without the finding in the group with the highest FPR than in the group with the lowest.

The null hypothesis is that all eligible groups have the same value of the measure. A test of this hypothesis has to account for the correlation between the images of one patient~\cite{Obuchowski1997Clustered,Genders2012Clustered}. So we resample patients. For each modality, we draw its test patients with replacement $B = 1{,}000$ times at seed 0. Each draw contains as many patients as the test set, and a drawn patient is included with all of their images, as often as the patient is drawn. The draws are not stratified by group or by label, and the same draws are used for all models, groups, measures, and strata of a modality. Patient identifiers are specific to a dataset. A lesion takes the place of the patient in ISIC 2019, and each image counts as one patient where a dataset releases no patient identifier.

Under the null hypothesis, the groups differ only through sampling variation. To obtain the distribution of the group difference in this case, we subtract from the value of each group in each resample its value on the full test set. Let $m^{*}_{jb}$ be the value of group $j$ in resample $b$. The recentered difference is
\begin{equation}\label{eq:chance}
\Delta^{\circ}_b = \max_{j} \left(m^{*}_{jb} - \hat m_j\right) - \min_{j} \left(m^{*}_{jb} - \hat m_j\right), \qquad b = 1, \dots, B ,
\end{equation}
where adding one common value to every recentered group value leaves the difference unchanged. The mean $\bar\Delta^{\circ}$ of the recentered differences is the difference expected by chance. Let $\mathcal B$ be the set of resamples in which the measure is defined for every eligible group. The AUROC needs images with and without the finding in every group, the sensitivity needs images with the finding, and the FPR needs images without the finding. Only the resamples of $\mathcal B$ are used. The one-sided $p$ value is
\begin{equation}\label{eq:p}
p = \frac{1 + \#\{b \in \mathcal B : \Delta^{\circ}_b \ge \hat\Delta\}}{1 + |\mathcal B|} ,
\end{equation}
where the addition of 1 prevents a $p$ value of zero. We apply the Benjamini-Hochberg false discovery rate (FDR) correction~\cite{BenjaminiHochberg1995}, write a corrected $p$ value as $p_{\mathrm{FDR}}$, and call a difference statistically significant when $p_{\mathrm{FDR}} < 0.05$. For the comparisons over all test patients, each modality, attribute, and measure defines one family of tests over all eligible models. These are the 416 chest radiograph models of both panels and 9 models in each of the other two modalities, the randomly initialized encoder included. The counts reported for one panel are parts of these families.

\subsection{Group differences within one hospital, radiograph view, and age band}

Over all test patients, the groups can differ in hospital, radiograph view, and age. To compare the groups of one hospital, we apply Eqs.~\ref{eq:obs} to~\ref{eq:p} to the test patients of this hospital alone, with the same patient resamples. For each hospital, attribute, and measure, the family of tests contains all eligible models.

When the view and the age band are held fixed as well, the groups of one stratum become small. So for these comparisons, we compute Eq.~\ref{eq:obs} within each stratum that has at least two eligible groups and test the mean over the $L$ strata,
\begin{equation}\label{eq:within}
\hat\Delta_{\mathrm{w}} = \frac{1}{L} \sum_{l=1}^{L} \hat\Delta_{l} , \qquad \Delta^{\circ}_{\mathrm{w},b} = \frac{1}{L} \sum_{l=1}^{L} \Delta^{\circ}_{l,b} ,
\end{equation}
where $\hat\Delta_l$ is the group difference within stratum $l$ and $\Delta^{\circ}_{l,b}$ is its recentered difference in resample $b$. We compare $\hat\Delta_{\mathrm{w}}$ with the $\Delta^{\circ}_{\mathrm{w},b}$ as in Eq.~\ref{eq:p}, over the resamples that are valid for every eligible group of every stratum. The stratifications are hospital, hospital and view, hospital and age band, and all three variables. Age differences are tested with hospital strata only, and race, sex, and insurance differences with all four stratifications. For each stratification, attribute, and measure, the family of tests contains all eligible chest radiograph models. The groups with the highest and the lowest value can differ between the strata of one model.

A difference that is not significant can still be large when the groups are small. So we report the critical value of each test,
\begin{equation}\label{eq:detect}
d = \bar\Delta^{\circ} + 1.645\, s\left(\Delta^{\circ}\right) ,
\end{equation}
where $s$ is the sample standard deviation (SD) of the recentered differences over the valid resamples and 1.645 is the 95th percentile of the standard normal distribution. Under a normal approximation of the null distribution, an observed difference above $d$ has a one-sided $p$ below 0.05 before the FDR correction. For the mean over strata, $d$ is computed from the $\Delta^{\circ}_{\mathrm{w},b}$. The decision of every test uses Eq.~\ref{eq:p}.

\subsection{Order of the groups at the shared threshold}

A difference in FPR alone does not show for which group a model performs worse, since a threshold that is low relative to the scores of a group raises both its FPR and its sensitivity. So for each chest radiograph model, we recorded the race group with the lowest FPR, the group with the highest FPR, and the group with the lowest sensitivity. We did this over all test patients, within MIMIC-CXR and within CheXpert, and within each age band of MIMIC-CXR, and for the two sex groups within MIMIC-CXR and within CheXpert. If the order of $k$ groups by FPR were unrelated to their order by sensitivity, the same group would have both lowest values with probability $1/k$. The sum of $1/k$ over the models is the number of models expected in this case.

We also computed the Spearman rank correlation between the FPR and the sensitivity across the race groups of each model. For each race group, we summarize each rate by its median over the models.

\subsection{Difference expected by chance from image counts}

Published studies report group values and group sizes, and not the scores and the patient identifiers that resampling needs. So we also define a count form of the test, which draws the null distribution from the group sizes and one common value $\bar m$ of the measure and treats every counted image as independent. We use it for the published studies and, in the ablation, in place of patient resampling for our models. For our models, $\bar m$ is the measure computed from the scores and labels of all eligible groups together. For each comparison, we draw $R = 2{,}000$ sets of group values at seed 0. Each group AUROC is a normal draw centered at $\bar m$, with the variance of Hanley and McNeil~\cite{HanleyMcNeil1982} at that value and at the numbers of images with and without the finding in the group, and it is clipped to $[0, 1]$. Sensitivity and FPR are binomial proportions with success probability $\bar m$, whose denominator is the number of images of the group with the finding for the sensitivity and without the finding for the FPR. The difference of draw $r$ is
\begin{equation}\label{eq:count}
\tilde\Delta_r = \max_{j} \tilde m_{jr} - \min_{j} \tilde m_{jr} , \qquad r = 1, \dots, R ,
\end{equation}
where $\tilde m_{jr}$ is the value of group $j$ in draw $r$. The mean $\bar{\tilde\Delta}$ is the difference expected by chance from image counts, and the one-sided $p$ value is computed as in Eq.~\ref{eq:p}, with the draws in place of the resamples. For our models, the count form uses the same families of tests as patient resampling, corrected separately.

For each model and attribute, the ratio $\bar\Delta^{\circ} / \bar{\tilde\Delta}$ compares the two differences expected by chance. For each measure, we summarize this ratio by its median over the race, sex, age, and insurance differences over all test patients. We report the median for the 117 chest radiograph models of the main panel and use the median over all 416 chest radiograph models for the published chest radiograph studies below.

\subsection{Group differences in published studies}

We assembled 22 studies that report the performance of a model by patient group, selected to cover the imaging modalities (Supplementary Note~\ref{snote:published}). A comparison required numerical group values of AUROC, sensitivity, or FPR, named groups, and their denominators, which for AUROC are the numbers of images with and without the finding. The counts were printed in the article or its supplement or were reconstructed from printed totals and group shares. Nine studies provided such comparisons (Supplementary Tables~\ref{stab:pubauroc} to~\ref{stab:pubrates}).

For each published comparison, $\bar m$ is the reported overall value of the measure or, where none is reported, the mean of the group values weighted by their denominators. We used Eq.~\ref{eq:count} to draw the null distribution and to compute the one-sided $p$ value. For AUROC, we used the Hanley-McNeil variance and, as a second form, the upper bound of Birnbaum and Klose on the variance of an AUROC~\cite{BirnbaumKlose1957}. For an AUROC $A$ from $n^{+}$ independent images with the finding and $n^{-}$ without it, the bound is $A(1-A)/\min(n^{+}, n^{-})$, which holds for every continuous distribution of the scores. We set $A = \bar m$ and drew normal values with this variance, clipped to $[0, 1]$.

The four chest radiograph studies report counts of images. For these studies, we also multiplied each difference drawn from the counts by the median ratio of patient resampling to image counts over our 416 chest radiograph models and the four attributes, i.e., by 1.148 for AUROC, 1.441 for sensitivity, and 2.144 for FPR. For each form of the test, the FDR correction is applied over the available comparisons of models and of dermatologists, with one family for AUROC and one for sensitivity and FPR together. Supplementary Note~\ref{snote:published} describes simulations of the count form of the AUROC test.

\subsection{Ablation of patient resampling}

To measure how often each form of the test is significant for groups that differ only by chance, we permuted the recorded race among the test patients with a recorded race. A permutation keeps the number of patients of each race group, and all images of a patient receive the same race. We used 100 permutations with seeds $1, 1{,}001, \dots, 99{,}001$ and tested all 13 findings for RAD-DINO and DINOv2 ViT-L in each of them. These two encoders were chosen before this analysis, as one chest radiograph encoder and one general encoder. We report the share of tests with an unadjusted $p$ below 0.05 for patient resampling and for the count form. Its SD and 95\% confidence interval (CI) come from 1{,}000 bootstrap resamples at seed 0 of the 1{,}300 pairs of permutation and finding, which keep both encoders of a pair together.

To measure how the test responds to a difference between the groups, we used five permutations and the nine public chest radiograph encoders plus the randomly initialized encoder. In each permutation and finding, we added independent Gaussian noise with mean zero to the scores of the eligible race group with the fewest images. The SD of the noise was 0, 0.5, or 1.0 times the SD of the scores of all images with a recorded race. The thresholds were not changed, and the scores with noise were not clipped. The SD and the CI of each share come from 1{,}000 bootstrap resamples at seed 0 of the 65 pairs of permutation and finding.

\subsection{Statistical analysis}

For a single model, a reported value of a measure or of a group difference is the estimate on the full test set. Its uncertainty is given as estimate $\pm$ SD (95\% CI), where the SD and the 2.5th and 97.5th percentiles come from the valid patient resamples. For a group difference, these resamples are not recentered, i.e., the uncertainty uses $\max_j m^{*}_{jb} - \min_j m^{*}_{jb}$, and the recentered differences of Eq.~\ref{eq:chance} are used for the test alone.

The tests of group differences are one-sided, since a group difference cannot be negative. The paired comparisons of pretraining runs in Supplementary Note~\ref{snote:replication} use two-sided tests of the change between two runs, centered at the observed change and paired by the shared patient resamples. Their families of tests contain the findings, backbones, and seeds of each comparison, attribute, and statistic. For the tests without RAD-DINO and the TorchXRayVision DenseNet-121, we repeated the FDR correction within each family over the remaining 390 chest radiograph models, of which 91 use public encoders.

A significant difference means that the groups differ in a measure by more than sampling variation explains. It does not identify the cause of the difference, and it does not mean that the model is unfair. A difference that is not significant is not evidence of equal performance, which is why we report the critical value of the test beside it. The models of a panel share their test patients, and several share an encoder, so that a count of models with a significant difference summarizes the panel and does not count independent replications.

\section{Results}

\subsection{Group differences in AUROC, sensitivity, and false positive rate}

\subsubsection{All test patients}

\begin{figure*}[!tbp]
\centering
\includegraphics[width=\textwidth]{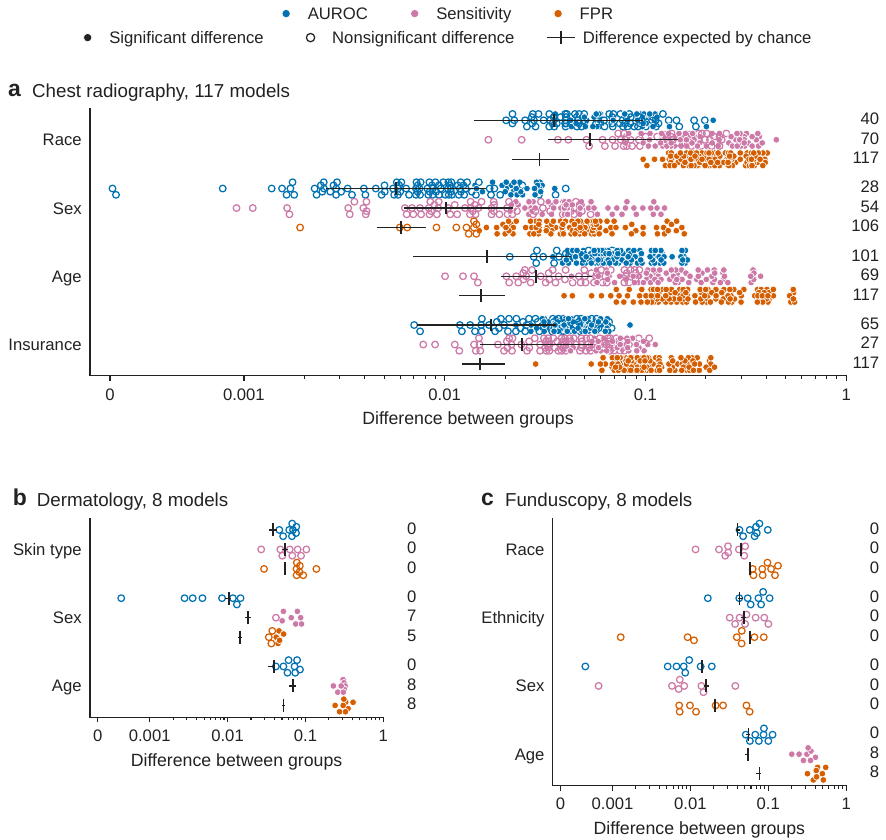}
\caption{Group differences of the models in three imaging modalities. Each dot is the difference between the largest and the smallest group value of one model, and a filled dot marks a statistically significant difference at a false discovery rate (FDR) of 0.05. The number at the right of each row counts its filled dots. A black line spans the smallest to the largest difference expected by chance across the models of a row, with a tick at the median. The horizontal axis is linear below 0.001 and logarithmic above.}
\label{fig:chance}
\end{figure*}

To answer RQ1, we first tested the group differences over all test patients of each modality. On chest radiographs, the race difference in FPR was statistically significant in all 117 models, and the race difference in AUROC in 40 (Fig.~\ref{fig:chance}a). Across models, the median race difference in FPR was 0.210, and the median difference expected by chance was 0.030. So in the median model, the race group with the highest FPR had 21 more false positive calls per 100 radiographs without the finding than the group with the lowest FPR. The median race difference in AUROC was 0.059, and the median difference expected by chance was 0.035. Supplementary Fig.~\ref{sfig:models} shows the race differences of every encoder and finding.

For example, the RAD-DINO model for pneumothorax called 34 of 100 radiographs of Asian patients without pneumothorax positive, compared with 11 of 100 radiographs of Hispanic or Latino patients. Its race difference in FPR was 0.231 $\pm$ 0.023 (0.184--0.275), and the difference expected by chance was 0.026 ($p_{\mathrm{FDR}} = 0.001$). Its race difference in AUROC was 0.041 $\pm$ 0.026 (0.017--0.112), with $p_{\mathrm{FDR}} = 0.544$.

The sex difference was significant in FPR in 106 of the 117 models and in AUROC in 28. Age and insurance differences in FPR were significant in all 117 models. In AUROC, they were significant in 101 models for age and in 65 for insurance. Sensitivity differences were significant in 70 models for race, 54 for sex, 69 for age, and 27 for insurance.

In dermatology and funduscopy, none of the 56 AUROC differences was significant (Fig.~\ref{fig:chance}b,c). The median critical value of the test was 0.074 for the AUROC difference by skin type, 0.077 by race on retinal images, and 0.097 by ethnicity. No difference by skin type, race, or ethnicity was significant in any measure. On retinal images, no sex difference was significant either. In dermatology, sex differences were significant in sensitivity in seven of the eight models and in FPR in five. Age differences in sensitivity and FPR were significant in all eight models of each modality, with a median FPR difference of 0.319 on dermatology images and of 0.423 on retinal images. Across the three modalities, 234 of the 524 AUROC differences and 478 of the 524 FPR differences were significant. On chest radiographs, these comparisons combine the patients of all hospitals.

\subsubsection{Race and sex within one hospital}

We then tested the race differences in each of the two hospitals separately. In MIMIC-CXR, no race difference in AUROC was significant in any of the 117 models. The median race difference in AUROC was 0.060, the median difference expected by chance was 0.046, and the median critical value was 0.082. In CheXpert, no race difference in AUROC was significant either, with a median difference of 0.034, a median difference expected by chance of 0.033, and a median critical value of 0.065. No race difference in sensitivity was significant in either hospital. By contrast, the race difference in FPR was significant in 107 of the 117 models in MIMIC-CXR and in 54 in CheXpert. Its median was 0.118 in MIMIC-CXR, where the median difference expected by chance was 0.038, and 0.059 in CheXpert, where it was 0.029. So within one hospital, the race groups with the highest and the lowest FPR differed by a median of 12 false positive calls per 100 radiographs without the finding in MIMIC-CXR and of 6 in CheXpert.

No sex difference in AUROC was significant in any model in MIMIC-CXR, CheXpert, NIH ChestX-ray14, or VinDr-CXR. In PadChest, the sex difference in AUROC was significant in 8 of the 54 models, with $p_{\mathrm{FDR}}$ from 0.016 to 0.043. Seven of these eight models detect cardiomegaly, with a higher AUROC for female patients and a difference of 0.023 to 0.036, and one detects consolidation. The sex difference in FPR was significant in 93 of the 117 models in MIMIC-CXR, in 55 of the 117 in CheXpert, in 46 of the 54 in PadChest, in 36 of the 90 in VinDr-CXR, and in none of the 54 in NIH ChestX-ray14.

\begin{figure*}[!tbp]
\centering
\includegraphics[width=\textwidth]{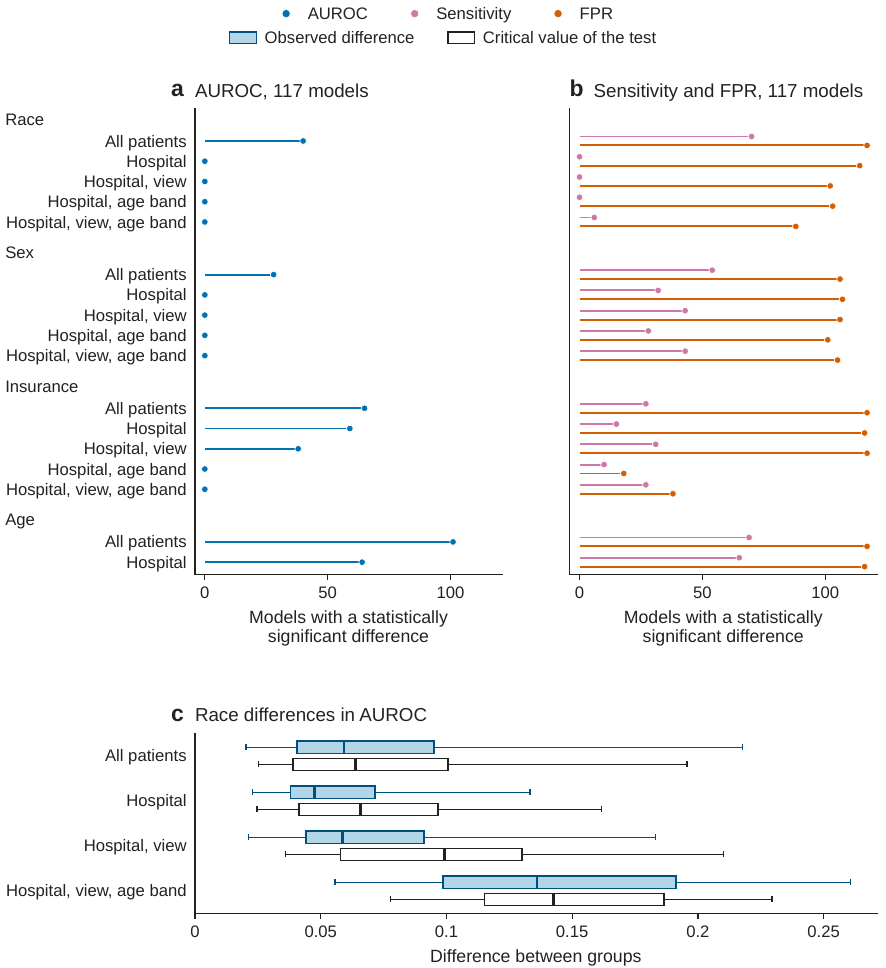}
\caption{Group differences of the 117 chest radiograph models over all patients and within strata. (\textbf{a} and \textbf{b}) Number of models with a statistically significant difference at an FDR of 0.05. In the rows below All patients, the difference is computed within each stratum of the named variables, and the mean over the strata is tested. Age is compared within hospitals only. (\textbf{c}) Race differences in AUROC and the critical value of the test (Eq.~\ref{eq:detect}). Boxes span the 25th to the 75th percentile of the 117 models, with a line at the median and whiskers to the smallest and the largest value.}
\label{fig:hospital}
\end{figure*}

Radiograph view and age can also differ between the groups of one hospital. So we computed each difference within the strata of hospital and view, of hospital and age band, and of all three variables, and tested the mean over the strata (Eq.~\ref{eq:within}). In every stratification, no mean race or sex difference in AUROC was significant in any of the 117 models (Fig.~\ref{fig:hospital}a). The mean race difference in FPR was significant in 114 models with hospital strata, in 102 with hospital and view strata, in 103 with hospital and age strata, and in 88 with all three variables (Fig.~\ref{fig:hospital}b). The mean sex difference in FPR was significant in 101 to 107 models in each stratification. With hospital strata, the median of the mean race differences in FPR was 0.092, and the median difference expected by chance was 0.035.

The median critical value for the mean race difference in AUROC was 0.066 with hospital strata, 0.099 with hospital and view strata, and 0.143 with all three variables (Fig.~\ref{fig:hospital}c). With all three variables, a median of 258 of the 1{,}000 resamples contained radiographs with and without the finding in every group of every stratum and were used in the test. With hospital strata, the median critical value for the mean sex difference in AUROC was 0.035. The mean race difference in sensitivity was significant in at most 6 models in each stratification, and the mean sex difference in sensitivity in 28 to 43.

\subsubsection{Age and insurance}

Age and insurance differences remained significant within hospitals. The mean age difference within hospitals was significant in AUROC in 64 of the 117 models and in FPR in 116 (Fig.~\ref{fig:hospital}a,b). The mean insurance difference within hospitals was significant in AUROC in 59 models and in FPR in 116. With hospital and view strata, it was significant in AUROC in 38 models and in FPR in 117.

The insurance groups differ in age. Patients aged 60 years or older contributed 81\% of the radiographs of patients with Medicare, compared with 28\% for Medicaid and 32\% for private insurance. So we also compared the insurance groups within the age bands of each hospital. With hospital and age strata, with or without the view, no mean insurance difference in AUROC was significant, at a median critical value of 0.107. The mean insurance difference in FPR was significant in 18 models with hospital and age strata and in 38 when the view was added. Its median decreased from 0.089 with hospital strata, where the median difference expected by chance was 0.020, to 0.063 with hospital and age strata, where it was 0.054. So within hospitals, most of the insurance differences were no longer significant once the age band was held fixed.

\subsection{Sensitivity and false positive rate of the groups at the shared threshold}

\begin{figure*}[!tbp]
\centering
\includegraphics[width=\textwidth]{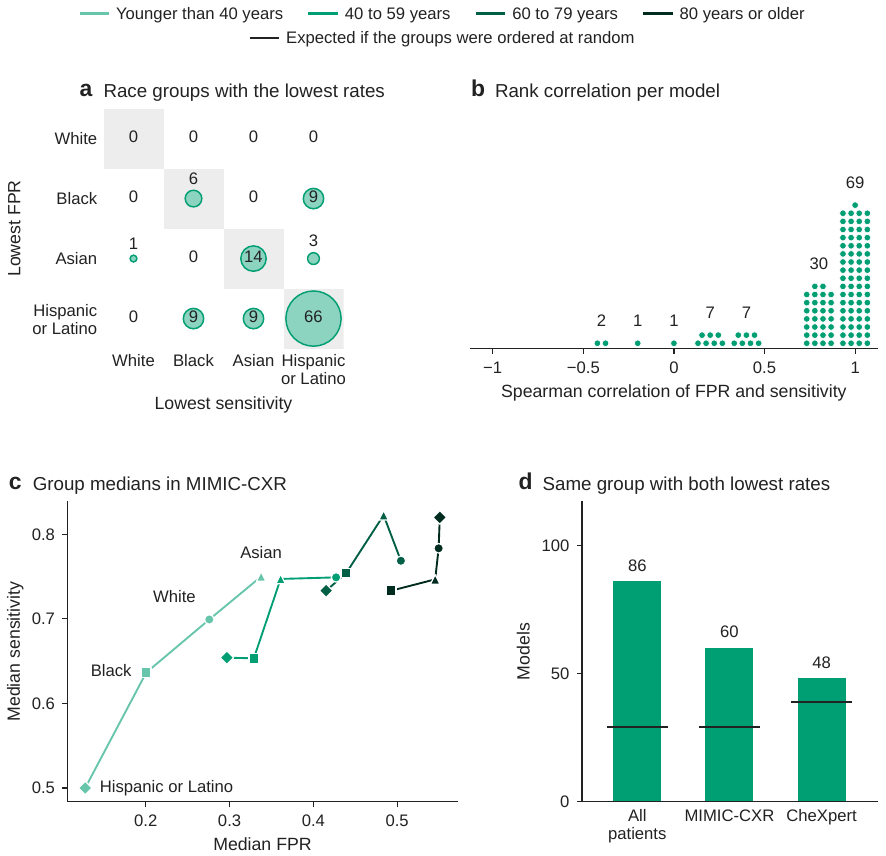}
\caption{Sensitivity and FPR of the race groups at the shared threshold in the 117 chest radiograph models. (\textbf{a}) Number of models by the race group with the lowest FPR and the race group with the lowest sensitivity over all patients. Bubble area is proportional to the number, and gray squares mark the cells in which one group has both lowest rates. (\textbf{b}) Spearman correlation of the FPR and the sensitivity across the four race groups over all patients, one dot per model. (\textbf{c}) Median FPR and median sensitivity of each race group in each age band of MIMIC-CXR~\cite{Johnson2019MIMICCXR}. Marker shapes distinguish the race groups, and lines join the groups of one age band. (\textbf{d}) Number of models in which one race group has both lowest rates, with three race groups in CheXpert~\cite{Irvin2019CheXpert} and four otherwise.}
\label{fig:threshold}
\end{figure*}

A lower FPR alone does not mean that a model performs better for a group. So we examined whether the group with the lowest FPR at the shared threshold also had the lowest sensitivity. In MIMIC-CXR, which has all four compared race groups, the same group had both lowest rates in 60 of the 117 models, compared with about 29 expected if the groups were ordered at random, independently by FPR and by sensitivity (Fig.~\ref{fig:threshold}d). FPR and sensitivity had a positive rank correlation across the race groups in 107 of the 117 models. The Hispanic or Latino group had the lowest FPR in 89 models and the lowest sensitivity in 61, and the White group had the highest FPR in 75. In CheXpert, which has three of the four race groups, the same group had both lowest rates in 48 models, compared with 39 expected. FPR and sensitivity had a positive rank correlation in 72 of these models. There, the Black group had the lowest FPR in 68 models and the lowest sensitivity in 69. Over all patients, the same group had both lowest rates in 86 models (Fig.~\ref{fig:threshold}a), and the rank correlation was positive in 113 (Fig.~\ref{fig:threshold}b).

We repeated the comparison within the age bands of MIMIC-CXR. In each of the three age bands below 80 years, the Hispanic or Latino group had the lowest FPR in 70 to 82 of the 117 models. In the band of 80 years or older, the Black group had the lowest FPR in 74 models. Among patients younger than 40 years, the models called a median of 13 of 100 radiographs without the finding positive for Hispanic or Latino patients and 28 of 100 for White patients. They detected the finding in a median of 50 of 100 radiographs with the finding for Hispanic or Latino patients and in 70 of 100 for White patients (Fig.~\ref{fig:threshold}c).

For sex, the group with the lower FPR also had the lower sensitivity in 102 of the 117 models in MIMIC-CXR and in 87 of the 117 in CheXpert, compared with about 59 expected for unrelated orders. All of these comparisons use the single threshold of each model, which was set at a validation sensitivity of 0.80.

\subsection{Encoders and their pretraining}

To answer RQ2, we compared the results between the encoders. For each of the nine public chest radiograph encoders, the mean race difference in FPR within hospitals was significant for 12 or 13 of the 13 findings. No race difference in AUROC was significant for any of these encoders in either hospital.

We then repeated the tests for the 299 models from the 22 encoders pretrained in this study and from the randomly initialized encoder (Supplementary Fig.~\ref{sfig:replication}). No race difference in AUROC was significant in any of the 299 models in either hospital. By contrast, the race difference in FPR was significant in 259 models in MIMIC-CXR and in 140 in CheXpert. The mean race difference in FPR within hospitals was significant in 146 of the 156 models from encoders pretrained on the natural data (median difference 0.089) and in 75 of the 78 from encoders pretrained on the race-balanced data (0.096). It was significant in 25 of the 26 models pretrained on reports without demographic terms (0.092) and in all 26 pretrained on reports that name the race, sex, and age band of the patient (0.102). For the randomly initialized encoder, it was significant in all 13 models (0.093). In dermatology and funduscopy, the models of the randomly initialized encoder had significant age differences in sensitivity and FPR and no other significant difference. By objective, it was significant in 125 of the 130 models with the image-report objective, in 74 of the 78 with the self-supervised objective, and in 73 of the 78 with the label-supervised objective. In paired comparisons of the pretraining runs over all test patients, no change of the objective, the data, or the reports changed the race difference in AUROC significantly in any of the 364 pairs (Supplementary Note~\ref{snote:replication}).

The pretraining data of RAD-DINO and of the TorchXRayVision DenseNet-121 include test radiographs of this study. Without these two encoders, no mean race difference in AUROC was significant in any of the remaining 91 models in any stratification. The mean race difference in FPR was significant in 89 of the 91 models with hospital strata and in 71 to 83 with the finer strata. One mean sex difference in AUROC was significant, for the DINOv3 ViT-S model for lung lesion with hospital and view strata ($p_{\mathrm{FDR}} = 0.049$).

\subsection{Group differences in published studies}

To examine whether the same contrast appears outside our models, we applied the count form of the test to nine published studies from six imaging modalities (Supplementary Fig.~\ref{sfig:published} and Supplementary Table~\ref{stab:published}). Of the 34 differences in AUROC between the patient groups of models, 1 was significant with the Hanley-McNeil variance and none with the variance bound. Of the 49 differences in sensitivity or FPR, 36 were significant. Of the 21 differences from chest radiograph studies, 16 remained significant after the adjustment for repeated images. For the dermatologists of one study, neither of the two differences in AUROC was significant, and all four differences in sensitivity or FPR were. Supplementary Note~\ref{snote:published} describes the studies and the simulations of the count form.

\subsection{Ablation of patient resampling}

\begin{figure*}[!tbp]
\centering
\includegraphics[width=\textwidth]{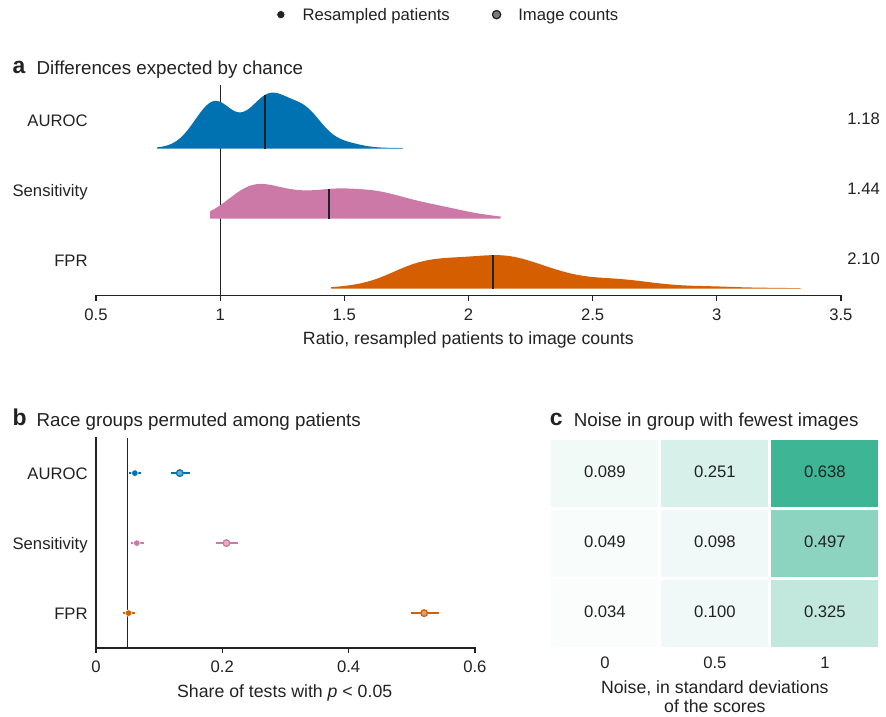}
\caption{Tests with resampled patients and with image counts. (\textbf{a}) Kernel density estimates of the difference expected by chance from resampled patients divided by that from image counts, over the 468 race, sex, age, and insurance differences of the 117 chest radiograph models. A black segment and the number at the right of each row mark the median. (\textbf{b}) Share of tests with $p < 0.05$ when the race groups are permuted among patients. Error bars span the 2.5th to the 97.5th percentile of 1{,}000 resamples, and the vertical line marks 0.05. (\textbf{c}) Share of tests with $p < 0.05$ for resampled patients when Gaussian noise is added to the scores of the permuted race group with the fewest images. The noise level is the standard deviation (SD) of the noise as a multiple of the SD of the scores. All $p$ values are unadjusted.}
\label{fig:images}
\end{figure*}

To measure the effect of resampling patients, we replaced it with the count form, which treats every image as independent, and kept the models, the groups, and the observed differences fixed. Over the race, sex, age, and insurance differences of the 117 chest radiograph models, the difference expected by chance from resampled patients was a median of 1.18 times that from image counts for AUROC, 1.44 times for sensitivity, and 2.10 times for FPR (Fig.~\ref{fig:images}a). It was the larger of the two in 356 of the 468 AUROC differences, in 464 of the 468 sensitivity differences, and in all 468 FPR differences. Of the 1{,}404 chest radiograph differences of Fig.~\ref{fig:chance}a, 911 were significant with resampled patients and 1{,}036 with image counts.

We then permuted the race groups among patients, so that the groups differed only by chance (Fig.~\ref{fig:images}b). With resampled patients, the share of tests with an unadjusted $p$ below 0.05 was 0.062 $\pm$ 0.005 (0.052--0.072) for AUROC, 0.065 $\pm$ 0.005 (0.055--0.076) for sensitivity, and 0.052 $\pm$ 0.005 (0.042--0.061) for FPR. With image counts, it was 0.133 $\pm$ 0.007 (0.118--0.148) for AUROC, 0.207 $\pm$ 0.009 (0.190--0.225) for sensitivity, and 0.520 $\pm$ 0.011 (0.498--0.543) for FPR. So with image counts, about half of the FPR tests had $p < 0.05$ for groups that differed only by chance.

Last, we added Gaussian noise to the scores of the permuted race group with the fewest images (Fig.~\ref{fig:images}c). For AUROC, the share of tests with $p < 0.05$ with resampled patients increased from 0.089 $\pm$ 0.021 (0.052--0.135) without noise to 0.251 $\pm$ 0.042 (0.169--0.337) with noise of 0.5 times the SD of the scores. With noise of 1.0 times the SD, it was 0.638 $\pm$ 0.052 (0.532--0.739). The median AUROC difference between the groups was 0.042 at 0.5 times the SD and 0.078 at 1.0 times the SD. At 1.0 times the SD, the share was 0.497 $\pm$ 0.051 (0.395--0.591) for sensitivity and 0.325 $\pm$ 0.043 (0.237--0.412) for FPR.

\section{Discussion}

This study tested whether the race and sex differences of medical imaging models are differences in ranking performance or differences in the error rates at the shared decision threshold (RQ1), and whether they depend on the image encoder (RQ2). For RQ1, no race difference in AUROC was significant in any of the 117 chest radiograph models in either hospital with recorded race. By contrast, the race difference in FPR was significant in 107 models in one hospital, at a median of 12 more false positive calls per 100 radiographs without the finding, and in 54 models in the other, at a median of 6. The group with the lowest FPR also had the lowest sensitivity more often than expected for unrelated orders. The sex groups differed significantly in AUROC in one of five datasets only, and mostly for one finding. For RQ2, this pattern held for each of the nine public encoders and for a randomly initialized encoder. It also held for the 22 encoders pretrained in this study with three objectives, with race-balanced data, and with edited reports.

Similar AUROC at different error rates has been reported for single models. Glocker et al. observed it for a DenseNet-121 and two labels~\cite{Glocker2023Encoding}, and Sutariya and Petersen for one model and one label~\cite{Sutariya2025Meval}. Our results show that this is the general case in the tested models and not a property of those single models. It appeared for every encoder and in both hospitals, with patients as the unit of every test. The results also show what a test set that combines hospitals adds. Over all test patients, the race difference in AUROC was significant in 40 of the 117 models, and within either hospital in none. The race groups come from the two hospitals in different proportions, so that a race difference measured over a combined test set can be a difference between hospitals. Therefore, a fairness evaluation on data from several hospitals should usually report each hospital separately.

At the shared threshold, the error rates differed in the direction of a trade-off. The group with fewer false positives also had more missed findings, so that a lower FPR did not mean that the model performed better for this group. This is the pattern of groups whose scores differ by a shift (Fig.~\ref{fig:overview}), for which one threshold is a different operating point in each group. Methods that change the threshold address differences of this type without retraining the model. Hardt et al. derived group-specific thresholds that equalize the error rates~\cite{Hardt2016Equality}, Lotter reduced an underdiagnosis difference with thresholds for each radiograph view~\cite{Lotter2024Acquisition}, and FairREAD applies thresholds specific to each subgroup~\cite{Gao2026FairREAD}. By contrast, none of the varied pretraining choices removed the differences. The median race difference in FPR within hospitals was 0.096 for encoders pretrained on race-balanced data, 0.093 for a randomly initialized encoder, and 0.092 for the public encoders. Therefore, a developer or a hospital that finds a race difference in a fairness evaluation could first examine the decision threshold. We did not test a threshold method ourselves.

Age and insurance gave other results than race and sex. Age differences within hospitals were significant in AUROC in 64 of the 117 models and in FPR in 116, so for age, the ranking differed as well. The insurance differences in AUROC were not significant once the groups were compared within age bands, and most of the insurance differences in FPR were not either, since the insurance groups differ in age. Therefore, a group attribute that correlates with age should usually be evaluated within age bands. The only recurring difference in AUROC by sex was for cardiomegaly in PadChest, where the AUROC was significantly higher for female patients with seven of the nine encoders. So a ranking difference can be specific to one finding in one hospital.

Counting images instead of patients changed the result of the test. With race groups permuted among patients, the test with image counts had $p < 0.05$ in 0.520 of the FPR tests, and the test with resampled patients in 0.052. Methods for clustered diagnostic data account for the correlation between images of one patient~\cite{Obuchowski1997Clustered,Genders2012Clustered}.

\subsection{Limitations and future work}

Race is recorded in only two of the chest radiograph datasets, both from hospitals in the United States, since the other public datasets do not release it. However, the result agreed between the two hospitals, and the sex comparisons cover five datasets from three countries. Future work should repeat the comparison in hospitals of other countries that record race or ethnicity.

A difference that is not significant has not been shown to be absent. The groups are smallest in the finest strata, where a median of 258 of the 1{,}000 resamples could be used for the race difference in AUROC, and in dermatology and funduscopy, where the critical values were 0.074 to 0.097. However, we report the critical value beside every result of this type. In MIMIC-CXR, the median critical value of the race difference in AUROC was 0.082, and the median observed difference was 0.060. In CheXpert, these two values were 0.065 and 0.034. Future work should test the equivalence of the group AUROCs at a margin fixed in advance, in cohorts large enough for it.

Every model is a logistic regression classifier on frozen features with one threshold, set at a validation sensitivity of 0.80. We tested no method that changes the threshold. Frozen features allowed us to compare 36 encoders under one procedure. However, the result was the same for all of these encoders, which differ in architecture, pretraining objective, and pretraining data. Future work should repeat the comparison for finetuned models at clinically chosen thresholds and should test whether thresholds for each group remove the differences without harm to any group.

Two public encoders were pretrained on data that include our test radiographs. The released list of pretraining images of RAD-DINO contains 114{,}625 of the 125{,}992 chest radiograph test images. The TorchXRayVision DenseNet-121 was trained with labels on data that include the same four datasets. However, without these two encoders, no mean race difference in AUROC was significant. The mean race difference in FPR remained significant in 71 to 89 of the 91 models. Future work should evaluate encoders on hospitals that are absent from every pretraining set.

The analysis of published studies uses the reported group values and group sizes of nine studies, since their scores and patient identifiers are not public. The count form of the test is an approximation. In simulations with equal AUROC in every group, it was significant in up to 0.086 of the studies with the variance bound and in up to 0.314 with the Hanley-McNeil variance (Supplementary Note~\ref{snote:published}). However, no AUROC difference was significant with the variance bound, and the adjustment for repeated images left most of the differences in error rates significant. Future work should repeat this analysis with the scores of the original models.

\section{Conclusion}

Within one hospital, the race groups of 117 chest radiograph models from nine public encoders differed significantly in FPR at the shared decision threshold. Their AUROC did not differ significantly in any model. The sex groups differed significantly in FPR in four of the five datasets and in AUROC in one. The same held for 299 models from encoders pretrained in this study with different objectives and data and from a randomly initialized encoder. Across the race groups of a hospital, a lower FPR came with a lower sensitivity in most models. In dermatology and funduscopy, no difference in AUROC by skin type, race, ethnicity, or sex was significant either. For the models and hospitals examined here, these results suggest that a race or sex difference in a fairness evaluation is first a question of the decision threshold of each group and hospital, and only then a question of the model.

\section*{Ethics statement and declaration}

All data are public and were used under the conditions of each release. The approval and the consent procedure reported for each dataset are given in Supplementary Note~\ref{snote:datasets}. No new patient data were generated and no participants were recruited. Therefore, this secondary analysis required no further ethics approval or informed consent. No re-identification was attempted, and no image or record is redistributed.

\section*{Declaration of generative AI and AI-assisted technologies in the writing process}

During the preparation of this work, the authors used Claude Opus 5.5 and GPT-6 Astra for language polishing. After using this tool, the authors reviewed and edited the content as needed and take full responsibility for the content of the publication.

\section*{Funding}

STA is supported by the German Research Foundation (DFG; 585806541), the German Federal Ministry of Research, Technology and Space (ARISTOTLE, 01ZU2602B), and the Excellence Strategy of the German Federal Government, the L\"ander, and RWTH ERS (START\_526-26).

DT is supported by the German Federal Ministry of Research, Technology and Space (TRANSFORM LIVER, 031L0312C; DECIPHER-M, 01KD2420B), the DFG (515639690), and the European Union (Horizon Europe, ODELIA, GA 101057091; ERC Starting Grant SAGMA, GA 101222556).

The funders had no role in the design of the study, in the collection, analysis, and interpretation of the data, in the writing of the manuscript, or in the decision to submit it for publication.

\section*{Contribution statement}

The formal analysis was conducted by ML and STA. The original draft was written by ML and STA and edited by STA. ML developed the code. The experiments were performed by ML. The statistical analyses were performed by ML and STA. DT provided clinical expertise. ML, DT, AM, and STA provided technical expertise. All authors read the manuscript and agreed to the submission of this paper.

\section*{Declaration of competing interest}

ML is employed by Generali Deutschland Services GmbH, Germany, and is on the editorial board of European Radiology Experimental. DT received honoraria for lectures from Bayer, GE, Roche, AstraZeneca, and Philips and holds shares in StratifAI GmbH, Germany, and in Synagen GmbH, Germany. AM is an associate editor at IEEE Transactions on Medical Imaging. STA received honoraria for lectures from RWTH International Academy gGmbH, Germany and is on the editorial boards of Communications Medicine and of European Radiology Experimental, and on the trainee editorial board of Radiology: Artificial Intelligence. The authors declare no other competing financial or non-financial interests.

\section*{Acknowledgments}

Not applicable.

\section*{Data availability}

Every dataset of this study is an existing public release, and no image or derived record is redistributed.

\begingroup
\footnotesize
\sloppy
\textbf{Chest radiography:}
\begin{itemize}
  \item MIMIC-CXR~\cite{Johnson2019MIMICCXR}, available from PhysioNet under credentialed access: \url{https://physionet.org/content/mimic-cxr-jpg/}. Access requires a signed data use agreement and completion of the required human subjects training.
  \item The race and insurance fields of MIMIC-IV~\cite{Johnson2023MIMICIV}, available under the same terms: \url{https://physionet.org/content/mimiciv/}.
  \item The MIMIC-CXR reports of the image-report pretraining runs, available under the same terms: \url{https://physionet.org/content/mimic-cxr/}.
  \item CheXpert~\cite{Irvin2019CheXpert} and the race and insurance fields of CheXpert Plus~\cite{Chambon2024ChexpertPlus}, available from the Stanford AIMI portal after registration and acceptance of its research use agreement: \url{https://aimi.stanford.edu/}.
  \item NIH ChestX-ray14~\cite{Wang2017ChestXray14}, openly available: \url{https://nihcc.app.box.com/v/ChestXray-NIHCC}.
  \item PadChest~\cite{Bustos2020PadChest}, available from the BIMCV repository on request: \url{https://bimcv.cipf.es/bimcv-projects/padchest/}.
  \item VinDr-CXR~\cite{Nguyen2022VinDrCXR} and VinDr-PCXR~\cite{Pham2022VinDrPCXR}, available from PhysioNet under credentialed access: \url{https://physionet.org/content/vindr-cxr/} and \url{https://physionet.org/content/vindr-pcxr/}.
\end{itemize}
\textbf{Dermatology:}
\begin{itemize}
  \item ISIC 2019~\cite{Tschandl2018HAM10000,Combalia2019BCN20000}, openly available: \url{https://challenge.isic-archive.com/}.
  \item Fitzpatrick17k~\cite{Groh2021Fitzpatrick17k}, openly available: \url{https://github.com/mattgroh/fitzpatrick17k}.
  \item DDI~\cite{Daneshjou2022DDI}, available from the Stanford AIMI portal after registration and acceptance of its research use agreement: \url{https://aimi.stanford.edu/}.
\end{itemize}
\textbf{Funduscopy:}
\begin{itemize}
  \item Harvard-FairVision~\cite{Luo2024FairVision}, available on request: \url{https://ophai.hms.harvard.edu/datasets/}.
\end{itemize}
\endgroup

The reported values and the counts used in the analysis of the 9 published studies are available in their articles and supplements.

The analysis code is publicly available at \url{https://github.com/mahshadlotfinia/bias_origin}. The repository does not redistribute the model weights or the underlying datasets. Every public encoder evaluated in this study has open weights, and all encoders were run on site without any cloud service or third-party application programming interface. The DINOv3, DermLIP, and RETFound checkpoints require accepting their terms on Hugging Face. The 22 encoders pretrained in this study start from the DINOv3 ViT-S and ViT-B checkpoints, and the randomly initialized ViT-S/16 has no released checkpoint. All models were accessed and all experiments were run between June and November 2026. The URLs of the evaluated checkpoints are:

\begingroup
\footnotesize
\sloppy
\textbf{Medical encoders for chest radiography:}
\begin{itemize}
  \item RAD-DINO: \url{https://huggingface.co/microsoft/rad-dino}. Its released list of pretraining images is at \url{https://huggingface.co/microsoft/rad-dino/blob/main/training_images.csv}.
  \item BiomedCLIP: \url{https://huggingface.co/microsoft/BiomedCLIP-PubMedBERT_256-vit_base_patch16_224}.
  \item TorchXRayVision DenseNet-121 (densenet121-res224-all): \url{https://github.com/mlmed/torchxrayvision}.
\end{itemize}
\textbf{General image encoders:}
\begin{itemize}
  \item DINOv3 ViT-L, ViT-B, and ViT-S: \url{https://huggingface.co/facebook/dinov3-vitl16-pretrain-lvd1689m}, \url{https://huggingface.co/facebook/dinov3-vitb16-pretrain-lvd1689m}, and \url{https://huggingface.co/facebook/dinov3-vits16-pretrain-lvd1689m}.
  \item DINOv2 ViT-L: \url{https://huggingface.co/facebook/dinov2-large}.
  \item CLIP ViT-L/14: \url{https://huggingface.co/openai/clip-vit-large-patch14}.
  \item SigLIP 2 ViT-L: \url{https://huggingface.co/google/siglip2-large-patch16-512}.
\end{itemize}
\textbf{Dermatology and funduscopy encoders:}
\begin{itemize}
  \item MONET: \url{https://huggingface.co/chanwkim/monet}.
  \item DermLIP ViT-B/16: \url{https://huggingface.co/redlessone/DermLIP_ViT-B-16}.
  \item RETFound: \url{https://huggingface.co/iszt/RETFound_mae_meh}.
  \item FLAIR: \url{https://github.com/jusiro/FLAIR}.
\end{itemize}
\textbf{Text encoder of the image-report pretraining runs:}
\begin{itemize}
  \item Bio\_ClinicalBERT: \url{https://huggingface.co/emilyalsentzer/Bio_ClinicalBERT}.
\end{itemize}
\endgroup

The study uses Python with PyTorch, torchvision, Hugging Face Transformers, huggingface-hub, timm, OpenCLIP, TorchXRayVision, and the FLAIR package. Data handling and statistical procedures use NumPy, SciPy, pandas, and scikit-learn. Figures use Matplotlib. Image loading and configuration use Pillow and PyYAML, and tqdm reports progress. Encoder pretraining and feature extraction ran on NVIDIA L40S GPUs (48~GB each) with Intel Xeon Silver 4310 CPUs, and the other analyses ran on CPUs.

\bibliographystyle{splncs04}
\bibliography{bibliography}

@article{BenjaminiHochberg1995,
  title={Controlling the false discovery rate: a practical and powerful approach to multiple testing},
  author={Benjamini, Yoav and Hochberg, Yosef},
  journal={Journal of the Royal statistical society: series B (Methodological)},
  volume={57},
  number={1},
  pages={289--300},
  year={1995},
}

@article{Johnson2019MIMICCXR,
  title={MIMIC-CXR, a de-identified publicly available database of chest radiographs with free-text reports},
  author={Johnson, Alistair EW and Pollard, Tom J and Berkowitz, Seth J and Greenbaum, Nathaniel R and Lungren, Matthew P and Deng, Chih-ying and Mark, Roger G and Horng, Steven},
  journal={Scientific data},
  volume={6},
  number={1},
  pages={317},
  year={2019},
  publisher={Nature Publishing Group UK London}
}

@inproceedings{Irvin2019CheXpert,
  title={Chexpert: A large chest radiograph dataset with uncertainty labels and expert comparison},
  author={Irvin, Jeremy and Rajpurkar, Pranav and Ko, Michael and others},
  booktitle={Proceedings of the AAAI conference on artificial intelligence},
  volume={33},
  number={01},
  pages={590--597},
  year={2019}
}

@article{Bustos2020PadChest,
title = {PadChest: A large chest x-ray image dataset with multi-label annotated reports},
journal = {Medical Image Analysis},
volume = {66},
pages = {101797},
year = {2020},
issn = {1361-8415},
doi = {https://doi.org/10.1016/j.media.2020.101797},
author = {Aurelia Bustos and Antonio Pertusa and Jose-Maria Salinas and Maria {de la Iglesia-Vayá}},
}

@article{Nguyen2022VinDrCXR,
  title={VinDr-CXR: An open dataset of chest X-rays with radiologist’s annotations},
  author={Nguyen, Ha Q and Lam, Khanh and Le, Linh T and others},
  journal={Scientific Data},
  volume={9},
  number={1},
  pages={429},
  year={2022},
  publisher={Nature Publishing Group UK London}
}

@article{
Daneshjou2022DDI,
author = {Roxana Daneshjou  and Kailas Vodrahalli  and Roberto A. Novoa  and others},
title = {Disparities in dermatology AI performance on a diverse, curated clinical image set},
journal = {Science Advances},
volume = {8},
number = {32},
pages = {eabq6147},
year = {2022},
doi = {10.1126/sciadv.abq6147},
}

@article{Combalia2019BCN20000,
  title={Bcn20000: Dermoscopic lesions in the wild},
  author={Hern{\'a}ndez-P{\'e}rez, Carlos and Combalia, Marc and others},
  journal={Scientific data},
  volume={11},
  number={1},
  pages={641},
  year={2024},
  publisher={Nature Publishing Group UK London}
}

@misc{Luo2024FairVision,
      title={FairVision: Equitable Deep Learning for Eye Disease Screening via Fair Identity Scaling},
      author={Yan Luo and Muhammad Osama Khan and Yu Tian and Min Shi and Zehao Dou and Tobias Elze and Yi Fang and Mengyu Wang},
      year={2024},
      eprint={2310.02492},
      archivePrefix={arXiv},
      primaryClass={cs.CV},
      url={https://arxiv.org/abs/2310.02492},
}

@article{PerezGarcia2025RadDino,
  title={Exploring scalable medical image encoders beyond text supervision},
  author={P{\'e}rez-Garc{\'\i}a, Fernando and Sharma, Harshita and Bond-Taylor, Sam and Bouzid, Kenza and Salvatelli, Valentina and Ilse, Maximilian and Bannur, Shruthi and Castro, Daniel C and Schwaighofer, Anton and Lungren, Matthew P and others},
  journal={Nature Machine Intelligence},
  volume={7},
  number={1},
  pages={119--130},
  year={2025},
  publisher={Nature Publishing Group UK London}
}

@misc{Simeoni2025DINOv3,
      title={DINOv3},
      author={Oriane Siméoni and Huy V. Vo and Maximilian Seitzer and others},
      year={2025},
      eprint={2508.10104},
      archivePrefix={arXiv},
      primaryClass={cs.CV},
      url={https://arxiv.org/abs/2508.10104},
}

@article{
Oquab2024DINOv2,
title={{DINO}v2: Learning Robust Visual Features without Supervision},
author={Maxime Oquab and Timoth{\'e}e Darcet and Th{\'e}o Moutakanni and others},
journal={Transactions on Machine Learning Research},
issn={2835-8856},
year={2024},
url={https://openreview.net/forum?id=a68SUt6zFt},
}

@inproceedings{Radford2021CLIP,
  title={Learning transferable visual models from natural language supervision},
  author={Radford, Alec and others},
  booktitle={ICML},
  pages={8748--8763},
  year={2021},
  organization={PmLR}
}

@misc{Tschannen2025SigLIP2,
      title={SigLIP 2: Multilingual Vision-Language Encoders with Improved Semantic Understanding, Localization, and Dense Features},
      author={Michael Tschannen and Alexey Gritsenko and Xiao Wang and Muhammad Ferjad Naeem and Ibrahim Alabdulmohsin and Nikhil Parthasarathy and Talfan Evans and Lucas Beyer and Ye Xia and Basil Mustafa and Olivier Hénaff and Jeremiah Harmsen and Andreas Steiner and Xiaohua Zhai},
      year={2025},
      eprint={2502.14786},
      archivePrefix={arXiv},
      primaryClass={cs.CV},
      url={https://arxiv.org/abs/2502.14786},
}

@article{Zhou2023RETFound,
  title={A foundation model for generalizable disease detection from retinal images},
  author={Zhou, Yukun and Chia, Mark A and Wagner, Siegfried K and others},
  journal={Nature},
  volume={622},
  number={7981},
  pages={156--163},
  year={2023},
  publisher={Nature Publishing Group UK London}
}

@article{Silva2025FLAIR,
  title={A foundation language-image model of the retina (flair): Encoding expert knowledge in text supervision},
  author={Silva-Rodriguez, Julio and Chakor, Hadi and Kobbi, Riadh and Dolz, Jose and Ayed, Ismail Ben},
  journal={Medical Image Analysis},
  volume={99},
  pages={103357},
  year={2025},
  publisher={Elsevier}
}

@article{Kim2024MONET,
  title={Transparent medical image AI via an image--text foundation model grounded in medical literature},
  author={Kim, Chanwoo and Gadgil, Soham U and DeGrave, Alex J and Omiye, Jesutofunmi A and Cai, Zhuo Ran and Daneshjou, Roxana and Lee, Su-In},
  journal={Nature medicine},
  volume={30},
  number={4},
  pages={1154--1165},
  year={2024},
  publisher={Nature Publishing Group US New York}
}

@article{HanleyMcNeil1982,
  title={The meaning and use of the area under a receiver operating characteristic (ROC) curve.},
  author={Hanley, James A and McNeil, Barbara J},
  journal={Radiology},
  volume={143},
  number={1},
  pages={29--36},
  year={1982}
}

@INPROCEEDINGS {Wang2017ChestXray14,
author = {Wang, Xiaosong and Peng, Yifan and Lu, Le and Lu, Zhiyong and Bagheri, Mohammadhadi and Summers, Ronald M.},
booktitle = {2017 IEEE Conference on Computer Vision and Pattern Recognition (CVPR) },
title = {{ ChestX-Ray8: Hospital-Scale Chest X-Ray Database and Benchmarks on Weakly-Supervised Classification and Localization of Common Thorax Diseases }},
year = {2017},
volume = {},
ISSN = {1063-6919},
pages = {3462-3471},
doi = {10.1109/CVPR.2017.369},
publisher = {IEEE Computer Society},
month =Jul}

@article{Pham2022VinDrPCXR,
  title={PediCXR: an open, large-scale chest radiograph dataset for interpretation of common thoracic diseases in children},
  author={Pham, Hieu H and Nguyen, Ngoc H and Tran, Thanh T and Nguyen, Tuan NM and Nguyen, Ha Q},
  journal={Scientific data},
  volume={10},
  number={1},
  pages={240},
  year={2023},
  publisher={Nature Publishing Group UK London}
}

@inproceedings{Groh2021Fitzpatrick17k,
  title={Evaluating deep neural networks trained on clinical images in dermatology with the fitzpatrick 17k dataset},
  author={Groh, Matthew and Harris, Caleb and Soenksen, Luis and Lau, Felix and Han, Rachel and Kim, Aerin and Koochek, Arash and Badri, Omar},
  booktitle={2021 IEEE/CVF Conference on Computer Vision and Pattern Recognition Workshops (CVPRW)},
  pages={1820--1828},
  year={2021},
  organization={IEEE}
}

@article{Zhang2023BiomedCLIP,
  title={A multimodal biomedical foundation model trained from fifteen million image--text pairs},
  author={Zhang, Sheng and Xu, Yanbo and Usuyama, Naoto and Xu, Hanwen and Bagga, Jaspreet and Tinn, Robert and Preston, Sam and Rao, Rajesh and Wei, Mu and Valluri, Naveen and others},
  journal={Nejm Ai},
  volume={2},
  number={1},
  pages={AIoa2400640},
  year={2025},
  publisher={Massachusetts Medical Society}
}

@inproceedings{Cohen2022TorchXRayVision,
  title={TorchXRayVision: A library of chest X-ray datasets and models},
  author={Cohen, Joseph Paul and Viviano, Joseph D and Bertin, Paul and Morrison, Paul and Torabian, Parsa and Guarrera, Matteo and Lungren, Matthew P and Chaudhari, Akshay and Brooks, Rupert and Hashir, Mohammad and others},
  booktitle={International Conference on Medical Imaging with Deep Learning},
  pages={231--249},
  year={2022},
  organization={PMLR}
}

@inproceedings{
Dosovitskiy2021ViT,
title={An Image is Worth 16x16 Words: Transformers for Image Recognition at Scale},
author={Alexey Dosovitskiy and Lucas Beyer and Alexander Kolesnikov and Dirk Weissenborn and Xiaohua Zhai and Thomas Unterthiner and Mostafa Dehghani and Matthias Minderer and Georg Heigold and Sylvain Gelly and Jakob Uszkoreit and Neil Houlsby},
booktitle={ICLR 2021},
year={2021},
url={https://openreview.net/forum?id=YicbFdNTTy}
}

@INPROCEEDINGS{Yan2025DermLIP,
  author={Yan, Siyuan and Hu, Ming and Jiang, Yiwen and Li, Xieji and Fei, Hao and Tschandl, Philipp and Kittler, Harald and Ge, Zongyuan},
  booktitle={2025 IEEE/CVF International Conference on Computer Vision (ICCV)},
  title={Derm1M: A Million-Scale Vision-Language Dataset Aligned with Clinical Ontology Knowledge for Dermatology},
  year={2025},
  volume={},
  number={},
  pages={12681-12690},
  doi={10.1109/ICCV51701.2025.01178}
  }

@article{Lotter2024Acquisition,
  title={Acquisition parameters influence AI recognition of race in chest x-rays and mitigating these factors reduces underdiagnosis bias},
  author={Lotter, William},
  journal={Nature Communications},
  volume={15},
  number={1},
  pages={7465},
  year={2024},
  publisher={Nature Publishing Group UK London}
}

@article{SeyyedKalantari2021Underdiagnosis,
  title={Underdiagnosis bias of artificial intelligence algorithms applied to chest radiographs in under-served patient populations},
  author={Seyyed-Kalantari, Laleh and Zhang, Haoran and McDermott, Matthew BA and Chen, Irene Y and Ghassemi, Marzyeh},
  journal={Nature medicine},
  volume={27},
  number={12},
  pages={2176--2182},
  year={2021},
  publisher={Nature Publishing Group US New York}
}

@article{Larrazabal2020Gender,
  title={Gender imbalance in medical imaging datasets produces biased classifiers for computer-aided diagnosis},
  author={Larrazabal, Agostina J and Nieto, Nicol{\'a}s and Peterson, Victoria and Milone, Diego H and Ferrante, Enzo},
  journal={Proceedings of the National Academy of Sciences},
  volume={117},
  number={23},
  pages={12592--12594},
  year={2020},
  publisher={National Academy of Sciences}
}

@article{Vaidya2024Pathology,
  title={Demographic bias in misdiagnosis by computational pathology models},
  author={Vaidya, Anurag and Chen, Richard J and Williamson, Drew FK and Song, Andrew H and Jaume, Guillaume and Yang, Yuzhe and Hartvigsen, Thomas and Dyer, Emma C and Lu, Ming Y and Lipkova, Jana and others},
  journal={Nature Medicine},
  volume={30},
  number={4},
  pages={1174--1190},
  year={2024},
  publisher={Nature Publishing Group US New York}
}

@article{Chen2023Algorithmic,
  title={Algorithmic fairness in artificial intelligence for medicine and healthcare},
  author={Chen, Richard J and Wang, Judy J and Williamson, Drew FK and Chen, Tiffany Y and Lipkova, Jana and Lu, Ming Y and Sahai, Sharifa and Mahmood, Faisal},
  journal={Nature biomedical engineering},
  volume={7},
  number={6},
  pages={719--742},
  year={2023},
  publisher={Nature Publishing Group UK London}
}

@inproceedings{
Zong2023MEDFAIR,
title={{MEDFAIR}: Benchmarking Fairness for Medical Imaging},
author={Yongshuo Zong and Yongxin Yang and Timothy Hospedales},
booktitle={The Eleventh International Conference on Learning Representations },
year={2023},
url={https://openreview.net/forum?id=6ve2CkeQe5S}
}

@article{Glocker2023Encoding,
  title={Algorithmic encoding of protected characteristics in chest X-ray disease detection models},
  author={Glocker, Ben and Jones, Charles and Bernhardt, M{\'e}lanie and Winzeck, Stefan},
  journal={EBioMedicine},
  volume={89},
  year={2023},
  publisher={Elsevier}
}

@article{Bernhardt2022Potential,
  title={Potential sources of dataset bias complicate investigation of underdiagnosis by machine learning algorithms},
  author={Bernhardt, M{\'e}lanie and Jones, Charles and Glocker, Ben},
  journal={Nature Medicine},
  volume={28},
  number={6},
  pages={1157--1158},
  year={2022},
  publisher={Nature Publishing Group US New York}
}

@article{Mukherjee2022Confounding,
  title={Confounding factors need to be accounted for in assessing bias by machine learning algorithms},
  author={Mukherjee, Pritam and Shen, Thomas C and Liu, Jianfei and Mathai, Tejas and Shafaat, Omid and Summers, Ronald M},
  journal={Nature Medicine},
  volume={28},
  number={6},
  pages={1159--1160},
  year={2022},
  publisher={Nature Publishing Group US New York}
}

@article{Hardt2016Equality,
  title={Equality of opportunity in supervised learning},
  author={Hardt, Moritz and Price, Eric and Srebro, Nati},
  journal={Advances in neural information processing systems},
  volume={29},
  year={2016}
}

@article{Pedregosa2011Sklearn,
  title={Scikit-learn: Machine learning in Python},
  author={Pedregosa, Fabian and Varoquaux, Ga{\"e}l and Gramfort, Alexandre and Michel, Vincent and Thirion, Bertrand and Grisel, Olivier and Blondel, Mathieu and Prettenhofer, Peter and Weiss, Ron and Dubourg, Vincent and others},
  journal={the Journal of machine Learning research},
  volume={12},
  pages={2825--2830},
  year={2011},
  publisher={JMLR. org}
}

@inproceedings{Alsentzer2019ClinicalBERT,
  title={Publicly available clinical BERT embeddings},
  author={Alsentzer, Emily and Murphy, John and Boag, William and Weng, Wei-Hung and Jindi, Di and Naumann, Tristan and McDermott, Matthew},
  booktitle={Proceedings of the 2nd clinical natural language processing workshop},
  pages={72--78},
  year={2019}
}

@inproceedings{Chen2020SimCLR,
  title={A simple framework for contrastive learning of visual representations},
  author={Chen, Ting and Kornblith, Simon and Norouzi, Mohammad and Hinton, Geoffrey},
  booktitle={International conference on machine learning},
  pages={1597--1607},
  year={2020},
  organization={PmLR}
}

@article{Tschandl2018HAM10000,
  title={The HAM10000 dataset, a large collection of multi-source dermatoscopic images of common pigmented skin lesions},
  author={Tschandl, Philipp and Rosendahl, Cliff and Kittler, Harald},
  journal={Scientific data},
  volume={5},
  number={1},
  pages={180161},
  year={2018},
  publisher={Nature Publishing Group}
}

@inproceedings{
Loshchilov2019AdamW,
title={Decoupled Weight Decay Regularization},
author={Ilya Loshchilov and Frank Hutter},
booktitle={International Conference on Learning Representations},
year={2019},
url={https://openreview.net/forum?id=Bkg6RiCqY7},
}

@article{TayebiArasteh2024Privacy,
  title={Preserving fairness and diagnostic accuracy in private large-scale AI models for medical imaging},
  author={Tayebi Arasteh, Soroosh and Ziller, Alexander and Kuhl, Christiane and Makowski, Marcus and Nebelung, Sven and Braren, Rickmer and Rueckert, Daniel and Truhn, Daniel and Kaissis, Georgios},
  journal={Communications medicine},
  volume={4},
  number={1},
  pages={46},
  year={2024},
  publisher={Nature Publishing Group UK London}
}

@article{Yang2024Limits,
  title={The limits of fair medical imaging AI in real-world generalization},
  author={Yang, Yuzhe and Zhang, Haoran and Gichoya, Judy W and Katabi, Dina and Ghassemi, Marzyeh},
  journal={Nature medicine},
  volume={30},
  number={10},
  pages={2838--2848},
  year={2024},
  publisher={Nature Publishing Group US New York}
}

@article{Johnson2023MIMICIV,
  title={MIMIC-IV, a freely accessible electronic health record dataset},
  author={Johnson, Alistair EW and Bulgarelli, Lucas and Shen, Lu and Gayles, Alvin and Shammout, Ayad and Horng, Steven and Pollard, Tom J and Hao, Sicheng and Moody, Benjamin and Gow, Brian and others},
  journal={Scientific data},
  volume={10},
  number={1},
  pages={1},
  year={2023},
  publisher={Nature Publishing Group UK London}
}

@article{Burlina2021Retinal,
  title={Addressing artificial intelligence bias in retinal diagnostics},
  author={Burlina, Philippe and Joshi, Neil and Paul, William and Pacheco, Katia D and Bressler, Neil M},
  journal={Translational Vision Science \& Technology},
  volume={10},
  number={2},
  pages={13--13},
  year={2021},
  publisher={The Association for Research in Vision and Ophthalmology}
}

@article{Yala2021Mirai,
  title={Multi-institutional validation of a mammography-based breast cancer risk model},
  author={Yala, Adam and Mikhael, Peter G and Strand, Fredrik and Lin, Gigin and Satuluru, Siddharth and Kim, Thomas and Banerjee, Imon and Gichoya, Judy and Trivedi, Hari and Lehman, Constance D and others},
  journal={Journal of Clinical Oncology},
  volume={40},
  number={16},
  pages={1732--1740},
  year={2022},
  publisher={Wolters Kluwer Health}
}

@article{Collins2024TRIPODAI,
  title={TRIPOD+ AI statement: updated guidance for reporting clinical prediction models that use regression or machine learning methods},
  author={Collins, Gary S and Moons, Karel GM and Dhiman, Paula and Riley, Richard D and Beam, Andrew L and Van Calster, Ben and Ghassemi, Marzyeh and Liu, Xiaoxuan and Reitsma, Johannes B and Van Smeden, Maarten and others},
  journal={bmj},
  volume={385},
  year={2024},
  publisher={British Medical Journal Publishing Group}
}

@misc{Chambon2024ChexpertPlus,
      title={CheXpert Plus: Augmenting a Large Chest X-ray Dataset with Text Radiology Reports, Patient Demographics and Additional Image Formats}, 
      author={Pierre Chambon and Jean-Benoit Delbrouck and Thomas Sounack and Shih-Cheng Huang and Zhihong Chen and Maya Varma and Steven QH Truong and Chu The Chuong and Curtis P. Langlotz},
      year={2024},
      eprint={2405.19538},
      archivePrefix={arXiv},
      primaryClass={cs.CL},
      url={https://arxiv.org/abs/2405.19538}, 
}

@inproceedings{Lum2022Debiasing,
  title={De-biasing “bias” measurement},
  author={Lum, Kristian and Zhang, Yunfeng and Bower, Amanda},
  booktitle={Proceedings of the 2022 ACM conference on fairness, accountability, and transparency},
  pages={379--389},
  year={2022}
}

@article{BirnbaumKlose1957,
  title={Bounds for the variance of the Mann-Whitney statistic},
  author={Birnbaum, Zygmund William and Klose, Orval M},
  journal={The Annals of Mathematical Statistics},
  pages={933--945},
  year={1957},
  publisher={JSTOR}
}

@article{Lekadir2025FUTUREAI,
  title={FUTURE-AI: international consensus guideline for trustworthy and deployable artificial intelligence in healthcare},
  author={Lekadir, Karim and Frangi, Alejandro F and Porras, Antonio R and Glocker, Ben and Cintas, Celia and Langlotz, Curtis P and Weicken, Eva and Asselbergs, Folkert W and Prior, Fred and Collins, Gary S and others},
  journal={bmj},
  volume={388},
  year={2025},
  publisher={British Medical Journal Publishing Group}
}

@article{TayebiArasteh2026Resolution,
  title={Resolution-dependent self-supervised transfer in chest radiograph classification},
  author={Tayebi Arasteh, Soroosh and Shaigan, Mina and Kuhl, Christiane and Kather, Jakob Nikolas and Nebelung, Sven and Truhn, Daniel},
  journal={Communications medicine},
  volume={6},
  number={1},
  pages={479},
  year={2026},
  publisher={Nature Publishing Group UK, London}
}

@article{Obuchowski1997Clustered,
  title={Nonparametric analysis of clustered ROC curve data},
  author={Obuchowski, Nancy A},
  journal={Biometrics},
  pages={567--578},
  year={1997},
  publisher={JSTOR}
}

@article{Genders2012Clustered,
  title={Methods for calculating sensitivity and specificity of clustered data: a tutorial},
  author={Genders, Tessa SS and Spronk, Sandra and Stijnen, Theo and Steyerberg, Ewout W and Lesaffre, Emmanuel and Hunink, MG Myriam},
  journal={Radiology},
  volume={265},
  number={3},
  pages={910--916},
  year={2012},
  publisher={Radiological Society of North America, Inc.}
}

@inproceedings{Sutariya2025Meval,
  title={meval: A Statistical Toolbox for Fine-Grained Model Performance Analysis},
  author={Sutariya, Dishantkumar and Petersen, Eike},
  booktitle={MICCAI Workshop on Fairness of AI in Medical Imaging},
  pages={187--197},
  year={2025},
  organization={Springer}
}

@article{TayebiArasteh2024Enhancing,
  title={Enhancing diagnostic deep learning via self-supervised pretraining on large-scale, unlabeled non-medical images},
  author={Tayebi Arasteh, Soroosh and Misera, Leo and Kather, Jakob Nikolas and Truhn, Daniel and Nebelung, Sven},
  journal={European Radiology Experimental},
  volume={8},
  number={1},
  pages={10},
  year={2024},
  publisher={Springer}
}

@article{Brown2023ShorT,
  title={Detecting shortcut learning for fair medical AI using shortcut testing},
  author={Brown, Alexander and Tomasev, Nenad and Freyberg, Jan and Liu, Yuan and Karthikesalingam, Alan and Schrouff, Jessica},
  journal={Nature communications},
  volume={14},
  number={1},
  pages={4314},
  year={2023},
  publisher={Nature Publishing Group UK, London}
}

@article{Gao2026FairREAD,
  title={FairREAD: Re-fusing demographic attributes after disentanglement for fair medical image classification},
  author={Gao, Yicheng and Hao, Jinkui and Zhou, Bo},
  journal={Medical Image Analysis},
  pages={103858},
  year={2025},
  publisher={Elsevier}
}

\clearpage

\setcounter{table}{0}
\setcounter{figure}{0}
\setcounter{equation}{0}
\renewcommand{\tablename}{Supplementary Table}
\renewcommand{\figurename}{Supplementary Fig.}
\floatname{algorithm}{Supplementary Algorithm}
\renewcommand{\thealgorithm}{\arabic{algorithm}}
\renewcommand{\theequation}{S\arabic{equation}}
\newcounter{suppnote}

\section*{Supplementary information}

\refstepcounter{suppnote}
\section*{Supplementary Note \thesuppnote: Datasets}
\label{snote:datasets}

\paragraph{Chest radiography.} We included frontal radiographs only and confirmed that each image existed before including its record. For NIH ChestX-ray14~\cite{Wang2017ChestXray14}, VinDr-CXR~\cite{Nguyen2022VinDrCXR}, and VinDr-PCXR~\cite{Pham2022VinDrPCXR}, we used the released test sets. The validation set of NIH ChestX-ray14 came from its released training and validation set. We split MIMIC-CXR~\cite{Johnson2019MIMICCXR}, CheXpert~\cite{Irvin2019CheXpert}, and PadChest~\cite{Bustos2020PadChest} by patient, with 18\% to 20\% of each dataset's images in the test set. Within each dataset, no recorded patient identifier appeared in more than one set. The dataset-specific identifiers could not rule out patient overlap between datasets. VinDr-CXR and VinDr-PCXR released no patient identifier.

The CheXpert metadata had already been filtered to remove 39{,}323 images with an uncertain label for at least one of cardiomegaly, lung opacity, lung lesion, edema, and pneumonia. We retained the 157{,}865 images recorded as anteroposterior or posteroanterior. We treated the recorded age of 238 years in one VinDr-CXR radiograph as missing. The five VinDr-CXR radiographs with recorded ages of 118 to 124 years were assigned to the oldest age band. Race and insurance strings were mapped to groups by case-insensitive matching. Unmatched strings were assigned to Other, a group that is left out of the comparisons. Sex values other than male and female, including the O of PadChest, were treated as missing.

\paragraph{Dermatology.} We retained the 16{,}012 images of Fitzpatrick17k~\cite{Groh2021Fitzpatrick17k} with a recorded skin type from its 16{,}577 images. The Diverse Dermatology Images (DDI) set~\cite{Daneshjou2022DDI} and Fitzpatrick17k released no patient identifier. We split their images individually. ISIC 2019~\cite{Tschandl2018HAM10000,Combalia2019BCN20000} was split by lesion. We used the released malignancy annotation for DDI and the 3-class label for Fitzpatrick17k. In ISIC 2019, melanoma, basal cell carcinoma, actinic keratosis, and squamous cell carcinoma were counted as malignant. Nevus, benign keratosis, dermatofibroma, and vascular lesion were counted as benign. Sex and age were recorded only for ISIC 2019. Skin type was recorded only for DDI and Fitzpatrick17k.

\paragraph{Funduscopy.} We used the released training, validation, and test sets of Harvard-FairVision~\cite{Luo2024FairVision} for glaucoma classification. Race was recorded as White, Black, or Asian. Ethnicity was recorded in a separate field.

\paragraph{Approvals.} MIMIC-CXR and MIMIC-IV~\cite{Johnson2023MIMICIV} were approved by the institutional review board of Beth Israel Deaconess Medical Center, which waived individual patient consent. CheXpert was approved by the institutional review board of Stanford Hospital, which waived consent for the de-identified data. CheXpert Plus~\cite{Chambon2024ChexpertPlus} adds de-identified reports and patient information from the same studies. PadChest was approved by an institutional research committee. VinDr-CXR was approved by the institutional review boards of Hanoi Medical University Hospital and Hospital 108. VinDr-PCXR was approved by the institutional review board of Phu Tho Obstetric and Pediatric Hospital. These boards waived individual patient consent.

The ethics committees of the Medical University of Vienna and the University of Queensland approved HAM10000~\cite{Tschandl2018HAM10000}. BCN20000~\cite{Combalia2019BCN20000} was collected with institutional ethics approval at Hospital Cl\'{\i}nic de Barcelona and with informed consent from every participant. DDI was approved by the Stanford Institutional Review Board. Harvard-FairVision was approved by the institutional review board of the institute that released it. Its data are de-identified.

\refstepcounter{suppnote}
\section*{Supplementary Note \thesuppnote: Paired comparisons of the pretraining runs}
\label{snote:replication}

We compared pairs of pretraining runs to measure how the race differences change with the pretraining objective, the pretraining data, and the reports. A comparison between two objectives pairs the runs with the same backbone, pretraining data, and seed. A comparison between the natural and the race-balanced data, or between the original and the edited reports, pairs the runs with the same objective and backbone at seed 0. For each run and finding, the race difference is the largest minus the smallest value of a measure across the recorded race groups over all test patients. We tested the change of this difference between the two runs of a pair with their 1{,}000 shared patient resamples, in a two-sided test with the resampled changes recentered at the observed change. In the same way, we tested the change of the overall area under the receiver operating characteristic curve (AUROC) and of the lowest AUROC among the recorded race groups. We applied the Benjamini-Hochberg false discovery rate (FDR) correction~\cite{BenjaminiHochberg1995} over the findings, backbones, and seeds of each comparison and statistic. A change is statistically significant when its corrected $p$ value, written as $p_{\mathrm{FDR}}$, is below 0.05.

In the 364 pairs, no change of the race difference in AUROC was significant (smallest $p_{\mathrm{FDR}} = 0.052$), and the race difference in sensitivity changed significantly in 3 pairs. On the natural data, the race difference in the false positive rate (FPR) changed significantly in 19 to 30 of the 52 pairs of each comparison between two objectives. Between two objectives, the absolute change of the race difference in FPR had a median of 0.041. Over the three seeds of one objective with the ViT-S backbone, the largest and the smallest race difference in FPR differed by a median of 0.013. Of the 54 significant changes between the image-report objective and one of the other two objectives, the race difference in FPR was smaller with the image-report objective in 51 and larger in 3. Between the natural and the race-balanced data, it changed significantly in 15 of the 78 pairs, with 7 increases and 8 decreases. Between the original and the edited reports, it changed significantly in none of the 52 pairs (smallest $p_{\mathrm{FDR}} = 0.104$).

On the natural data, the overall AUROC was significantly higher with the image-report objective than with the self-supervised objective in all 52 pairs, by a median of 0.043. It was also significantly higher than with the label-supervised objective in all 52 pairs, by a median of 0.048. With the image-report objective, the lowest AUROC among the race groups was significantly higher than with the self-supervised objective in 36 of the 52 pairs and than with the label-supervised objective in 40.

\refstepcounter{suppnote}
\section*{Supplementary Note \thesuppnote: Published studies}
\label{snote:published}

We examined 22 published studies, selected to cover the imaging modalities. Nine of them report numerical group values and enough information to obtain the group sizes. These nine studies cover six imaging modalities and provide 89 comparisons between patient groups, 83 of models and 6 of the dermatologists evaluated by Daneshjou et al.~\cite{Daneshjou2022DDI}. Supplementary Tables~\ref{stab:pubauroc} to~\ref{stab:pubrates} list the comparisons, and Supplementary Table~\ref{stab:published} summarizes the results by study. Of the other 13 studies, 5 report other statistics, 5 plot the group results without numerical values, and 3 give no group counts.

Where only a part of the results of a study was eligible, we analyzed this part. For example, we included the abdominal computed tomography results of Tayebi Arasteh et al.~\cite{TayebiArasteh2024Privacy} and not their chest radiograph AUROC, which is averaged over eight diagnoses. We used counts printed in captions and did not estimate values from plotted marks. The denominators are printed counts, differences between printed totals and counts, or products of printed totals and shares. Where a study does not print the counts of its groups, we reconstructed them from the reported composition of the dataset and the fraction of the test split. We kept the group definitions of each study and required at least two groups with a value and a denominator.

For the four chest radiograph studies, the Adjusted columns of the tables use the differences drawn from the image counts, multiplied by the median ratio between patient resampling and image counts in the 416 chest radiograph models of the present study. This ratio is 1.148 for AUROC, 1.441 for sensitivity, and 2.144 for FPR. The FDR correction was applied separately for each form of the test, with one family for AUROC and one for sensitivity and FPR together, over the available comparisons of models and of dermatologists.

\begin{figure*}[t]
\centering
\includegraphics[width=\textwidth]{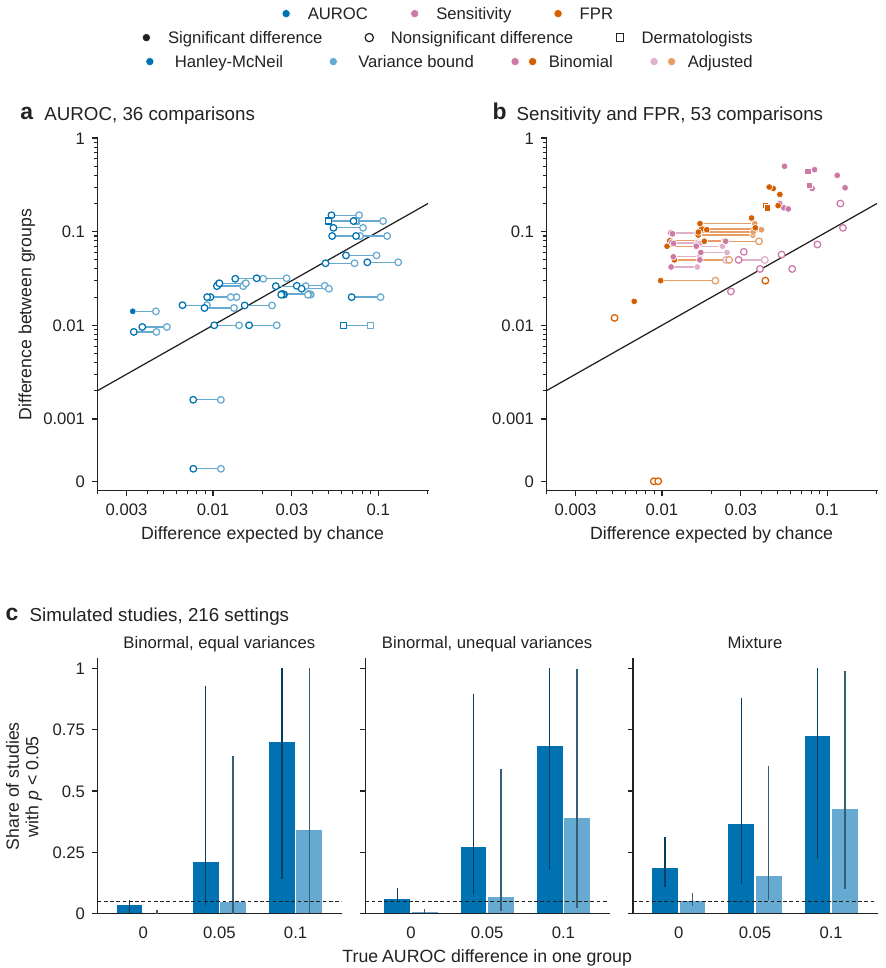}
\caption{Group differences in published studies and simulations of the count form of the test. (\textbf{a} and \textbf{b}) Each marker pairs the reported difference of one comparison with the difference expected by chance at its group sizes, and a line joins the markers of one comparison. For AUROC, the difference expected by chance uses the Hanley-McNeil variance~\cite{HanleyMcNeil1982} or the variance bound~\cite{BirnbaumKlose1957}. For sensitivity and FPR, it uses binomial draws, and the light markers show it after the adjustment for repeated images. A filled marker is a statistically significant difference at an FDR of 0.05. On the diagonal, the two differences are equal. (\textbf{c}) Median share of simulated studies with an unadjusted $p < 0.05$ over the 24 settings of each score distribution and AUROC difference. Lines span the smallest to the largest share, and the dashed line marks 0.05.}
\label{sfig:published}
\end{figure*}

Of the 34 differences in AUROC between the patient groups of models, 1 was significant with the Hanley-McNeil variance and none with the variance bound (Supplementary Fig.~\ref{sfig:published}a). For example, Glocker et al.~\cite{Glocker2023Encoding} reported race differences in AUROC of 0.01 or 0.02, and the differences expected by chance at their group sizes were 0.009 to 0.017 with the Hanley-McNeil variance.

Of the 49 differences in sensitivity or FPR between the patient groups of models, 36 were significant (Supplementary Fig.~\ref{sfig:published}b). Of the 21 differences from chest radiograph studies, 16 remained significant after the adjustment. For example, Seyyed-Kalantari et al.~\cite{SeyyedKalantari2021Underdiagnosis} reported the underdiagnosis rate, i.e., the FPR of the label No finding, whose denominator is the radiographs with at least one finding. This rate differed by 0.122 between the groups defined by race and sex together in MIMIC-CXR, and the difference expected by chance at their group sizes was 0.017. For the dermatologists, all four differences in sensitivity or FPR were significant, and neither of the two differences in AUROC was.

To test the count form of the AUROC test, we simulated 1{,}000 studies in each of 216 settings. The settings cross 2 or 4 groups, a baseline AUROC of 0.75 or 0.90, 25, 100, or 400 images with the finding per group, and 1 or 9 images without the finding per image with it. One group has an AUROC that is lower than the baseline by 0, 0.05, or 0.10. The three score distributions are binormal with equal variances, binormal with twice the standard deviation (SD) for the scores of images with the finding, and a mixture in which each score of an image with the finding lies above or below all scores of images without it. For the binormal distributions, the scores of images without the finding have mean 0 and SD 1. The scores of images with the finding have SD $\sigma = 1$ or 2 and mean $\sqrt{1 + \sigma^2}\,\Phi^{-1}(A)$, where $A$ is the AUROC of the group and $\Phi^{-1}$ is the standard normal quantile function. For the mixture, the scores of images without the finding are uniform on $[0, 1]$, and the scores of images with the finding are uniform on $[10, 11]$ with probability $A$ and on $[-10, -9]$ otherwise.

The settings are ordered by score distribution, baseline AUROC, AUROC difference, number of images with the finding, ratio of images without to images with the finding, and number of groups, with the last factor varying fastest. Each setting uses its index from 0 to 215 as the seed of its scores. The seed of the draws from the counts is the sum of the index of the setting and the index of the simulated study, which is between 0 and 999. The common AUROC of the count form is computed from all scores of images with the finding and at most the first 4{,}000 scores of images without it, in the order of the groups. We report the share of studies with an unadjusted $p$ below 0.05 for the Hanley-McNeil variance and for the variance bound.

In these simulations (Supplementary Fig.~\ref{sfig:published}c), the largest share of studies with an unadjusted $p < 0.05$ among the settings with the same AUROC in every group was 0.086 with the variance bound and 0.314 with the Hanley-McNeil variance. The largest share with the Hanley-McNeil variance occurred for the mixture distribution. In the 72 settings in which the AUROC of one group is lower by 0.10, the median share was 0.371 with the variance bound and 0.713 with the Hanley-McNeil variance.

\clearpage

\begin{figure*}[t]
\centering
\includegraphics[width=\textwidth]{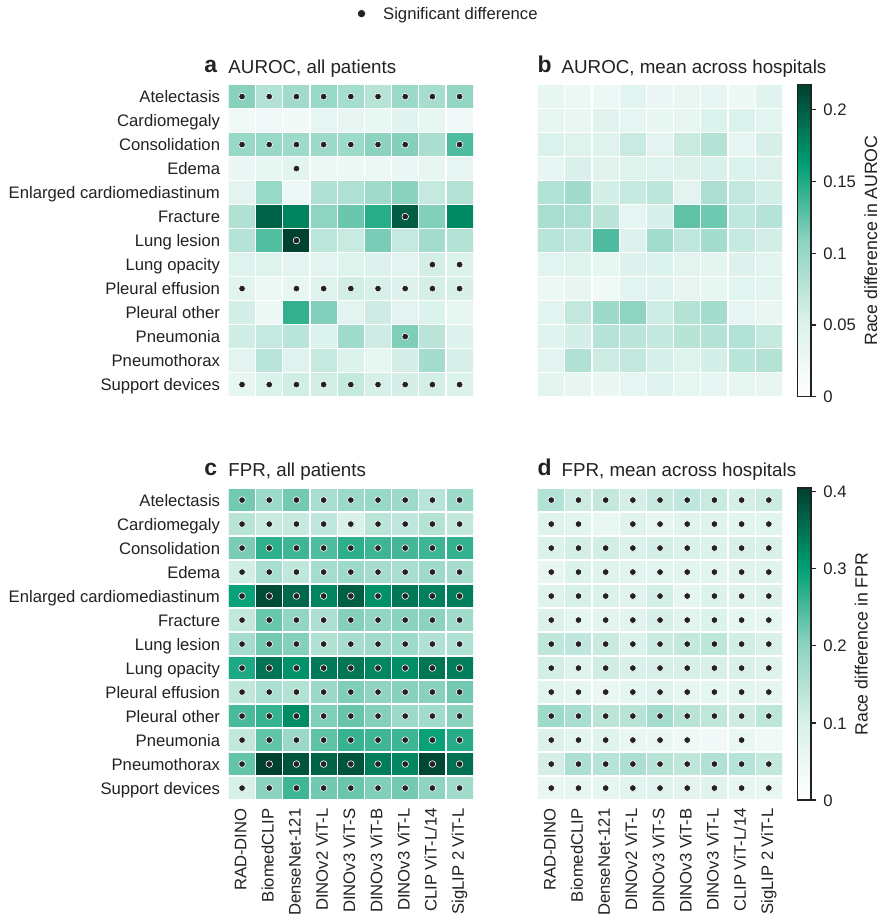}
\caption{Race differences of the 117 chest radiograph models by encoder and finding. The color shows the difference between the largest and the smallest value of the race groups. The encoders are RAD-DINO~\cite{PerezGarcia2025RadDino}, BiomedCLIP~\cite{Zhang2023BiomedCLIP}, the TorchXRayVision DenseNet-121~\cite{Cohen2022TorchXRayVision}, DINOv2 ViT-L~\cite{Oquab2024DINOv2}, DINOv3 ViT-S, ViT-B, and ViT-L~\cite{Simeoni2025DINOv3}, CLIP ViT-L/14~\cite{Radford2021CLIP}, and SigLIP 2 ViT-L~\cite{Tschannen2025SigLIP2}. Panels \textbf{a} and \textbf{c} use all test patients, and panels \textbf{b} and \textbf{d} show the mean of the differences within MIMIC-CXR~\cite{Johnson2019MIMICCXR} and within CheXpert~\cite{Irvin2019CheXpert}. A dot marks a statistically significant difference at a false discovery rate (FDR) of 0.05. AUROC, area under the receiver operating characteristic curve; FPR, false positive rate; ViT, vision transformer.}
\label{sfig:models}
\end{figure*}

\begin{figure*}[t]
\centering
\includegraphics[width=\textwidth]{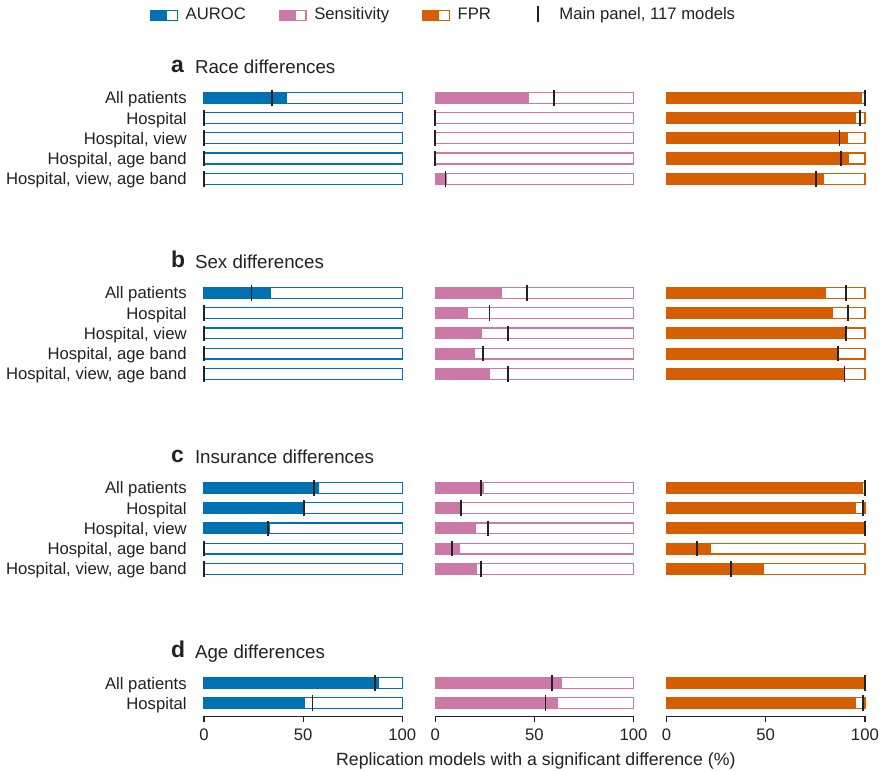}
\caption{Group differences of the replication panel over all patients and within strata. The replication panel has 299 chest radiograph models from 22 encoders pretrained in this study and from one randomly initialized ViT-S/16~\cite{Dosovitskiy2021ViT}. Each bar is filled up to the share of models with a statistically significant difference at an FDR of 0.05, and a black tick marks this share among the 117 models of the main panel. In the rows below All patients, the difference is computed within each stratum of the named variables, and the mean over the strata is tested. Age is compared within hospitals only.}
\label{sfig:replication}
\end{figure*}

\clearpage

\begin{table}[p]
\centering
\caption{Published differences in the area under the receiver operating characteristic curve (AUROC) compared with chance, one row per comparison. Rows are grouped by study, with its imaging modality, and the dermatologists of Daneshjou et al.~\cite{Daneshjou2022DDI} are listed after its models. DDI common is the subset of DDI with common diseases. The Groups column lists the number of compared groups and, in parentheses, the total number of cases in the smallest group, including cases with and without the finding, as counted by the study. The difference expected by chance is computed at the group sizes with the Hanley-McNeil variance or with the variance bound of Birnbaum and Klose~\cite{BirnbaumKlose1957}. Each $p_{\mathrm{FDR}}$ is a one-sided $p$ value after the FDR correction over all AUROC comparisons available for that form of the test. The correction in Adjusted includes only the chest radiograph comparisons. In that column, the chance distribution of the Hanley-McNeil test of a chest radiograph study is multiplied by 1.148. CT, computed tomography; DDI, Diverse Dermatology Images; ERM, empirical risk minimization; FDR, false discovery rate; N/A, not applicable; PDAC, pancreatic ductal adenocarcinoma.}
\label{stab:pubauroc}
\setlength{\tabcolsep}{2pt}
\renewcommand{\arraystretch}{1.0}
{\scriptsize
\begin{tabular*}{\textwidth}{@{\extracolsep{\fill}}>{\raggedright\arraybackslash}p{0.11\textwidth}>{\raggedright\arraybackslash}p{0.12\textwidth}>{\raggedright\arraybackslash}p{0.13\textwidth}llrrrrrr@{}}
\toprule
 & & & & & & \multicolumn{2}{c}{Hanley-McNeil} & \multicolumn{2}{c}{Variance bound} & Adjusted \\
\cmidrule(lr){7-8}\cmidrule(lr){9-10}\cmidrule(l){11-11}
Data & Task & Condition & Attribute & Groups & Difference & Chance & $p_{\mathrm{FDR}}$ & Chance & $p_{\mathrm{FDR}}$ & $p_{\mathrm{FDR}}$ \\
\midrule
\multicolumn{11}{@{}l}{\textbf{Glocker et al.}~\cite{Glocker2023Encoding}, chest radiography} \\
CheXpert & No finding & N/A & Race & 3 (2{,}746) & 0.010 & 0.017 & 0.864 & 0.024 & 0.975 & 0.985 \\
CheXpert & Pleural effusion & N/A & Race & 3 (2{,}746) & 0.020 & 0.009 & 0.157 & 0.013 & 0.559 & 0.205 \\
MIMIC-CXR & No finding & N/A & Race & 3 (2{,}082) & 0.010 & 0.010 & 0.707 & 0.014 & 0.975 & 0.843 \\
MIMIC-CXR & Pleural effusion & N/A & Race & 3 (2{,}082) & 0.020 & 0.010 & 0.157 & 0.014 & 0.584 & 0.205 \\
\midrule
\multicolumn{11}{@{}l}{\textbf{Zong et al.}~\cite{Zong2023MEDFAIR}, chest radiography} \\
CheXpert & No finding & ERM & Race & 2 (9{,}716) & 0.0016 & 0.008 & 0.924 & 0.011 & 0.975 & 0.985 \\
CheXpert & No finding & ERM & Sex & 2 (9{,}057) & 0.0002 & 0.008 & 0.982 & 0.011 & 0.986 & 0.985 \\
MIMIC-CXR & No finding & ERM & Race & 2 (14{,}630) & 0.0085 & 0.003 & 0.157 & 0.005 & 0.559 & 0.205 \\
MIMIC-CXR & No finding & ERM & Sex & 2 (17{,}743) & 0.0141 & 0.003 & 0.036 & 0.005 & 0.468 & 0.040 \\
\midrule
\multicolumn{11}{@{}l}{\textbf{Tayebi Arasteh et al.}~\cite{TayebiArasteh2024Privacy}, abdominal CT} \\
Test set & PDAC & $\varepsilon = 0.29$ & Age & 4 (75) & 0.0471 & 0.086 & 0.924 & 0.132 & 0.986 & N/A \\
Test set & PDAC & $\varepsilon = 0.54$ & Age & 4 (75) & 0.0558 & 0.064 & 0.827 & 0.098 & 0.975 & N/A \\
Test set & PDAC & $\varepsilon = 1.06$ & Age & 4 (75) & 0.046 & 0.048 & 0.781 & 0.072 & 0.975 & N/A \\
Test set & PDAC & $\varepsilon = 2.04$ & Age & 4 (75) & 0.0246 & 0.034 & 0.864 & 0.050 & 0.975 & N/A \\
Test set & PDAC & $\varepsilon = 4.71$ & Age & 4 (75) & 0.0213 & 0.027 & 0.827 & 0.039 & 0.975 & N/A \\
Test set & PDAC & $\varepsilon = 5.0$ & Age & 4 (75) & 0.0214 & 0.027 & 0.827 & 0.039 & 0.975 & N/A \\
Test set & PDAC & $\varepsilon = 6.0$ & Age & 4 (75) & 0.0216 & 0.026 & 0.827 & 0.038 & 0.975 & N/A \\
Test set & PDAC & $\varepsilon = 7.0$ & Age & 4 (75) & 0.0212 & 0.026 & 0.827 & 0.038 & 0.975 & N/A \\
Test set & PDAC & $\varepsilon = 8.0$ & Age & 4 (75) & 0.0163 & 0.016 & 0.707 & 0.023 & 0.975 & N/A \\
Test set & PDAC & Non-private & Age & 4 (75) & 0.0153 & 0.009 & 0.207 & 0.013 & 0.740 & N/A \\
Test set & PDAC & $\varepsilon = 0.29$ & Sex & 2 (127) & 0.0264 & 0.032 & 0.781 & 0.047 & 0.975 & N/A \\
Test set & PDAC & $\varepsilon = 0.54$ & Sex & 2 (127) & 0.0262 & 0.024 & 0.690 & 0.036 & 0.975 & N/A \\
Test set & PDAC & $\varepsilon = 1.06$ & Sex & 2 (127) & 0.0318 & 0.018 & 0.319 & 0.028 & 0.740 & N/A \\
Test set & PDAC & $\varepsilon = 2.04$ & Sex & 2 (127) & 0.0314 & 0.014 & 0.185 & 0.020 & 0.584 & N/A \\
Test set & PDAC & $\varepsilon = 4.71$ & Sex & 2 (127) & 0.0281 & 0.011 & 0.157 & 0.016 & 0.559 & N/A \\
Test set & PDAC & $\varepsilon = 5.0$ & Sex & 2 (127) & 0.0279 & 0.011 & 0.157 & 0.016 & 0.559 & N/A \\
Test set & PDAC & $\varepsilon = 6.0$ & Sex & 2 (127) & 0.0267 & 0.011 & 0.157 & 0.015 & 0.559 & N/A \\
Test set & PDAC & $\varepsilon = 7.0$ & Sex & 2 (127) & 0.0262 & 0.011 & 0.157 & 0.015 & 0.559 & N/A \\
Test set & PDAC & $\varepsilon = 8.0$ & Sex & 2 (127) & 0.0164 & 0.007 & 0.157 & 0.009 & 0.559 & N/A \\
Test set & PDAC & Non-private & Sex & 2 (127) & 0.0096 & 0.004 & 0.157 & 0.005 & 0.559 & N/A \\
\midrule
\multicolumn{11}{@{}l}{\textbf{Daneshjou et al.}~\cite{Daneshjou2022DDI}, models, dermatology} \\
DDI & Malignancy & ModelDerm & Skin tone & 2 (207) & 0.090 & 0.053 & 0.319 & 0.078 & 0.740 & N/A \\
DDI & Malignancy & DeepDerm & Skin tone & 2 (207) & 0.110 & 0.054 & 0.246 & 0.081 & 0.730 & N/A \\
DDI & Malignancy & HAM10000 model & Skin tone & 2 (207) & 0.150 & 0.052 & 0.157 & 0.077 & 0.559 & N/A \\
DDI common & Malignancy & ModelDerm & Skin tone & 2 (156) & 0.020 & 0.069 & 0.923 & 0.103 & 0.975 & N/A \\
DDI common & Malignancy & DeepDerm & Skin tone & 2 (156) & 0.090 & 0.073 & 0.609 & 0.113 & 0.975 & N/A \\
DDI common & Malignancy & HAM10000 model & Skin tone & 2 (156) & 0.130 & 0.071 & 0.307 & 0.107 & 0.740 & N/A \\
\midrule
\multicolumn{11}{@{}l}{\textbf{Daneshjou et al.}~\cite{Daneshjou2022DDI}, dermatologists, dermatology} \\
DDI & Malignancy & Ensemble & Skin tone & 2 (207) & 0.130 & 0.050 & 0.157 & 0.073 & 0.559 & N/A \\
DDI common & Malignancy & Ensemble & Skin tone & 2 (156) & 0.010 & 0.061 & 0.925 & 0.089 & 0.975 & N/A \\
\bottomrule
\end{tabular*}}
\end{table}

\begin{table}[p]
\centering
\caption{Published differences in sensitivity and false positive rate (FPR) of chest radiograph models compared with chance, one row per comparison. Rows are grouped by study, with the attribute of its compared groups. The Groups column lists the number of compared groups and, in parentheses, the number of cases of the smallest group, counted with the finding for sensitivity and without it for FPR. The difference expected by chance is computed at the group sizes from binomial draws. In the columns Adjusted, its chance distribution is multiplied by 1.441 for sensitivity and by 2.144 for FPR. The FDR correction combines sensitivity and FPR comparisons across modalities for Binomial and only chest radiograph comparisons for Adjusted. A difference is statistically significant when its $p_{\mathrm{FDR}}$ is below 0.05. The label Race and sex denotes groups defined by both attributes together. The group sizes of Lotter~\cite{Lotter2024Acquisition} were derived from printed values. For Lotter, the condition names the training data and the method.}
\label{stab:pubchest}
\setlength{\tabcolsep}{2pt}
\renewcommand{\arraystretch}{1.02}
{\scriptsize
\begin{tabular*}{\textwidth}{@{\extracolsep{\fill}}>{\raggedright\arraybackslash}p{0.12\textwidth}>{\raggedright\arraybackslash}p{0.11\textwidth}>{\raggedright\arraybackslash}p{0.2\textwidth}llrrrrr@{}}
\toprule
 & & & & & & \multicolumn{2}{c}{Binomial} & \multicolumn{2}{c}{Adjusted} \\
\cmidrule(lr){7-8}\cmidrule(l){9-10}
Data & Task & Condition & Measure & Groups & Difference & Chance & $p_{\mathrm{FDR}}$ & Chance & $p_{\mathrm{FDR}}$ \\
\midrule
\multicolumn{10}{@{}l}{\textbf{Seyyed-Kalantari et al.}~\cite{SeyyedKalantari2021Underdiagnosis}, chest radiography, race and sex} \\
MIMIC-CXR & No finding & N/A & FPR & 4 (1{,}152) & 0.122 & 0.017 & 0.001 & 0.036 & 0.003 \\
\midrule
\multicolumn{10}{@{}l}{\textbf{Glocker et al.}~\cite{Glocker2023Encoding}, chest radiography, race} \\
CheXpert & No finding & N/A & Sensitivity & 3 (312) & 0.050 & 0.029 & 0.136 & 0.042 & 0.320 \\
CheXpert & No finding & N/A & FPR & 3 (2{,}434) & 0.030 & 0.010 & 0.007 & 0.021 & 0.223 \\
CheXpert & Pleural effusion & N/A & Sensitivity & 3 (897) & 0.070 & 0.016 & 0.001 & 0.023 & 0.008 \\
CheXpert & Pleural effusion & N/A & FPR & 3 (1{,}849) & 0.050 & 0.012 & 0.001 & 0.025 & 0.071 \\
MIMIC-CXR & No finding & N/A & Sensitivity & 3 (659) & 0.060 & 0.017 & 0.002 & 0.025 & 0.031 \\
MIMIC-CXR & No finding & N/A & FPR & 3 (1{,}423) & 0.080 & 0.011 & 0.001 & 0.024 & 0.002 \\
MIMIC-CXR & Pleural effusion & N/A & Sensitivity & 3 (578) & 0.050 & 0.017 & 0.010 & 0.024 & 0.069 \\
MIMIC-CXR & Pleural effusion & N/A & FPR & 3 (1{,}504) & 0.070 & 0.011 & 0.001 & 0.023 & 0.015 \\
\midrule
\multicolumn{10}{@{}l}{\textbf{Lotter}~\cite{Lotter2024Acquisition}, chest radiography, race} \\
MIMIC-CXR & Any finding & MIMIC-CXR & Sensitivity & 3 (1{,}303) & 0.077 & 0.011 & 0.001 & 0.017 & 0.002 \\
MIMIC-CXR & Any finding & MIMIC-CXR, augmentation & Sensitivity & 3 (1{,}303) & 0.097 & 0.011 & 0.001 & 0.016 & 0.002 \\
MIMIC-CXR & Any finding & MIMIC-CXR, per-view threshold & Sensitivity & 3 (1{,}303) & 0.042 & 0.011 & 0.002 & 0.016 & 0.030 \\
MIMIC-CXR & Any finding & CheXpert & Sensitivity & 3 (1{,}303) & 0.075 & 0.012 & 0.001 & 0.017 & 0.002 \\
MIMIC-CXR & Any finding & CheXpert, augmentation & Sensitivity & 3 (1{,}303) & 0.095 & 0.012 & 0.001 & 0.017 & 0.002 \\
MIMIC-CXR & Any finding & CheXpert, per-view threshold & Sensitivity & 3 (1{,}303) & 0.054 & 0.012 & 0.001 & 0.017 & 0.008 \\
MIMIC-CXR & Any finding & MIMIC-CXR & FPR & 3 (689) & 0.092 & 0.017 & 0.001 & 0.036 & 0.031 \\
MIMIC-CXR & Any finding & MIMIC-CXR, augmentation & FPR & 3 (689) & 0.099 & 0.017 & 0.001 & 0.036 & 0.023 \\
MIMIC-CXR & Any finding & MIMIC-CXR, per-view threshold & FPR & 3 (689) & 0.079 & 0.018 & 0.001 & 0.039 & 0.076 \\
MIMIC-CXR & Any finding & CheXpert & FPR & 3 (689) & 0.109 & 0.018 & 0.001 & 0.038 & 0.020 \\
MIMIC-CXR & Any finding & CheXpert, augmentation & FPR & 3 (689) & 0.107 & 0.017 & 0.001 & 0.037 & 0.019 \\
MIMIC-CXR & Any finding & CheXpert, per-view threshold & FPR & 3 (689) & 0.105 & 0.019 & 0.001 & 0.040 & 0.030 \\
\bottomrule
\end{tabular*}}
\end{table}

\begin{table}[p]
\centering
\caption{Published differences in sensitivity and FPR in other imaging modalities compared with chance, one row per comparison. Rows are grouped by study, with its imaging modality and the attribute of its compared groups, and the dermatologists of Daneshjou et al.~\cite{Daneshjou2022DDI} are listed after its models. The columns have the same definitions as in Supplementary Table~\ref{stab:pubchest}. DR, diabetic retinopathy; IDC, invasive ductal carcinoma; IDH1, isocitrate dehydrogenase 1; ILC, invasive lobular carcinoma; LUAD, lung adenocarcinoma; LUSC, lung squamous cell carcinoma; MGB, Mass General Brigham; TCGA-GBM/LGG, The Cancer Genome Atlas glioblastoma and lower-grade glioma cohorts.}
\label{stab:pubrates}
\setlength{\tabcolsep}{2pt}
\renewcommand{\arraystretch}{1.02}
{\scriptsize
\begin{tabular*}{\textwidth}{@{\extracolsep{\fill}}>{\raggedright\arraybackslash}p{0.13\textwidth}>{\raggedright\arraybackslash}p{0.15\textwidth}>{\raggedright\arraybackslash}p{0.2\textwidth}llrrr@{}}
\toprule
 & & & & & & \multicolumn{2}{c}{Binomial} \\
\cmidrule(lr){7-8}
Data & Task & Condition & Measure & Groups & Difference & Chance & $p_{\mathrm{FDR}}$ \\
\midrule
\multicolumn{8}{@{}l}{\textbf{Yala et al.}~\cite{Yala2021Mirai}, mammography, race} \\
Emory & Cancer in 5 years & At Tyrer-Cuzick specificity & Sensitivity & 2 (301) & 0.061 & 0.031 & 0.147 \\
Emory & Cancer in 5 years & At Tyrer-Cuzick specificity & FPR & 2 (3{,}422) & 0.018 & 0.007 & 0.044 \\
Emory & Cancer in 5 years & At Tyrer-Cuzick sensitivity & Sensitivity & 2 (301) & 0.023 & 0.026 & 0.524 \\
Emory & Cancer in 5 years & At Tyrer-Cuzick sensitivity & FPR & 2 (3{,}422) & 0.012 & 0.005 & 0.086 \\
\midrule
\multicolumn{8}{@{}l}{\textbf{Vaidya et al.}~\cite{Vaidya2024Pathology}, pathology, race} \\
MGB breast & Breast subtype & IDC & Sensitivity & 3 (69) & 0.057 & 0.053 & 0.447 \\
MGB breast & Breast subtype & ILC & Sensitivity & 3 (55) & 0.175 & 0.058 & 0.004 \\
MGB lung & Lung subtype & LUAD & Sensitivity & 3 (100) & 0.079 & 0.024 & 0.006 \\
MGB lung & Lung subtype & LUSC & Sensitivity & 3 (26) & 0.073 & 0.087 & 0.602 \\
TCGA-GBM/LGG & IDH1 mutation & Wild-type & Sensitivity & 3 (9) & 0.110 & 0.125 & 0.524 \\
TCGA-GBM/LGG & IDH1 mutation & Mutant & Sensitivity & 3 (15) & 0.295 & 0.128 & 0.036 \\
\midrule
\multicolumn{8}{@{}l}{\textbf{Daneshjou et al.}~\cite{Daneshjou2022DDI}, models, dermatology, skin tone} \\
DDI & Malignancy & ModelDerm & Sensitivity & 2 (48) & 0.290 & 0.081 & 0.006 \\
DDI & Malignancy & ModelDerm & FPR & 2 (159) & 0.140 & 0.035 & 0.003 \\
DDI & Malignancy & DeepDerm & Sensitivity & 2 (48) & 0.460 & 0.084 & 0.001 \\
DDI & Malignancy & DeepDerm & FPR & 2 (159) & 0.300 & 0.045 & 0.001 \\
DDI & Malignancy & HAM10000 model & Sensitivity & 2 (48) & 0.040 & 0.039 & 0.524 \\
DDI & Malignancy & HAM10000 model & FPR & 2 (159) & 0.000 & 0.009 & $>$ 0.999 \\
DDI common & Malignancy & ModelDerm & Sensitivity & 2 (16) & 0.200 & 0.120 & 0.218 \\
DDI common & Malignancy & ModelDerm & FPR & 2 (140) & 0.110 & 0.037 & 0.020 \\
DDI common & Malignancy & DeepDerm & Sensitivity & 2 (16) & 0.400 & 0.115 & 0.009 \\
DDI common & Malignancy & DeepDerm & FPR & 2 (140) & 0.290 & 0.047 & 0.001 \\
DDI common & Malignancy & HAM10000 model & Sensitivity & 2 (16) & 0.040 & 0.061 & 0.628 \\
DDI common & Malignancy & HAM10000 model & FPR & 2 (140) & 0.000 & 0.009 & $>$ 0.999 \\
\midrule
\multicolumn{8}{@{}l}{\textbf{Daneshjou et al.}~\cite{Daneshjou2022DDI}, dermatologists, dermatology, skin tone} \\
DDI & Malignancy & Ensemble & Sensitivity & 2 (48) & 0.440 & 0.076 & 0.001 \\
DDI & Malignancy & Ensemble & FPR & 2 (159) & 0.190 & 0.042 & 0.001 \\
DDI common & Malignancy & Ensemble & Sensitivity & 2 (16) & 0.310 & 0.078 & 0.004 \\
DDI common & Malignancy & Ensemble & FPR & 2 (140) & 0.180 & 0.044 & 0.003 \\
\midrule
\multicolumn{8}{@{}l}{\textbf{Burlina et al.}~\cite{Burlina2021Retinal}, funduscopy, skin tone} \\
Test set & Referable DR & Baseline & Sensitivity & 2 (100) & 0.500 & 0.055 & 0.001 \\
Test set & Referable DR & Baseline & FPR & 2 (100) & 0.250 & 0.052 & 0.001 \\
Test set & Referable DR & Debiased, retina & Sensitivity & 2 (100) & 0.180 & 0.055 & 0.017 \\
Test set & Referable DR & Debiased, retina & FPR & 2 (100) & 0.030 & 0.042 & 0.665 \\
Test set & Referable DR & Debiased, DR status & Sensitivity & 2 (100) & 0.200 & 0.052 & 0.004 \\
Test set & Referable DR & Debiased, DR status & FPR & 2 (100) & 0.190 & 0.050 & 0.005 \\
\bottomrule
\end{tabular*}}
\end{table}

\clearpage

\begin{table}[p]
\centering
\caption{Group differences of published studies, tested from their reported group values and group sizes. Each cell is the number of statistically significant differences among the comparisons of one study and attribute. The one-sided tests use 2{,}000 draws from the group sizes. A difference is statistically significant when its $p_{\mathrm{FDR}}$ is below 0.05, corrected separately for each form of the test and for AUROC and for sensitivity and FPR together. The Adjusted columns apply the adjustment for repeated images and cover the chest radiograph studies only. The label Race and sex denotes groups defined by both attributes together. The bold row sums the comparisons of models, without the dermatologists of Daneshjou et al.}
\label{stab:published}
\setlength{\tabcolsep}{4pt}
\renewcommand{\arraystretch}{1.08}
{\footnotesize
\begin{tabular*}{\textwidth}{@{\extracolsep{\fill}}>{\raggedright\arraybackslash}p{0.3\textwidth}lrrrrr@{}}
\toprule
 & & \multicolumn{3}{c}{AUROC} & \multicolumn{2}{c}{Sensitivity and FPR} \\
\cmidrule(lr){3-5}\cmidrule(l){6-7}
Study & Attribute & \shortstack[r]{Hanley-\\McNeil} & \shortstack[r]{Variance\\bound} & Adjusted & Binomial & Adjusted \\
\midrule
\multicolumn{7}{@{}l}{\textbf{Chest radiography}} \\
Seyyed-Kalantari et al.~\cite{SeyyedKalantari2021Underdiagnosis} & Race and sex & N/A & N/A & N/A & 1 of 1 & 1 of 1 \\
Glocker et al.~\cite{Glocker2023Encoding} & Race & 0 of 4 & 0 of 4 & 0 of 4 & 7 of 8 & 4 of 8 \\
Lotter~\cite{Lotter2024Acquisition} & Race & N/A & N/A & N/A & 12 of 12 & 11 of 12 \\
Zong et al.~\cite{Zong2023MEDFAIR} & Race, sex & 1 of 4 & 0 of 4 & 1 of 4 & N/A & N/A \\
\midrule
\multicolumn{7}{@{}l}{\textbf{Mammography}} \\
Yala et al.~\cite{Yala2021Mirai} & Race & N/A & N/A & N/A & 1 of 4 & N/A \\
\midrule
\multicolumn{7}{@{}l}{\textbf{Abdominal CT}} \\
Tayebi Arasteh et al.~\cite{TayebiArasteh2024Privacy} & Age, sex & 0 of 20 & 0 of 20 & N/A & N/A & N/A \\
\midrule
\multicolumn{7}{@{}l}{\textbf{Pathology}} \\
Vaidya et al.~\cite{Vaidya2024Pathology} & Race & N/A & N/A & N/A & 3 of 6 & N/A \\
\midrule
\multicolumn{7}{@{}l}{\textbf{Dermatology}} \\
Daneshjou et al.~\cite{Daneshjou2022DDI}, models & Skin tone & 0 of 6 & 0 of 6 & N/A & 7 of 12 & N/A \\
Daneshjou et al.~\cite{Daneshjou2022DDI}, dermatologists & Skin tone & 0 of 2 & 0 of 2 & N/A & 4 of 4 & N/A \\
\midrule
\multicolumn{7}{@{}l}{\textbf{Funduscopy}} \\
Burlina et al.~\cite{Burlina2021Retinal} & Skin tone & N/A & N/A & N/A & 5 of 6 & N/A \\
\midrule
\textbf{All models} & & \textbf{1 of 34} & \textbf{0 of 34} & \textbf{1 of 8} & \textbf{36 of 49} & \textbf{16 of 21} \\
\bottomrule
\end{tabular*}}
\end{table}

\end{document}